\documentclass[acmsmall]{acmart}
\usepackage{multirow}
\usepackage{makecell}
\usepackage{subfigure}
\usepackage{upgreek}
\usepackage{threeparttable}
\usepackage{bbding}
\usepackage{pifont}
\usepackage{algorithmic}
\usepackage{algorithm}

\AtBeginDocument{%
  }

\setcopyright{acmlicensed}
\acmJournal{TKDD}
\acmYear{2025} \acmVolume{1} \acmNumber{1} \acmArticle{1} \acmMonth{1}\acmDOI{10.1145/3758099}

\acmJournal{TKDD}
\acmVolume{37}
\acmNumber{4}
\acmArticle{111}
\acmMonth{1}

\begin{document}

\title{Multi-Grained Spatial-Temporal Feature Complementarity for Accurate Online Cellular Traffic Prediction}

\author{Ningning Fu}
	\orcid{0009-0009-7407-6597}
\affiliation{%
  \institution{School of Information Science and Engineering, Southeast University}
  \city{Nanjing}
  \country{China}}
\email{funingning@seu.edu.cn}

\author{Shengheng Liu}
	\authornote{Corresponding author}
	\orcid{0000-0001-6579-9798}
\affiliation{%
  \institution{School of Information Science and Engineering, Southeast University}
  \city{Nanjing}
  \country{China}}
	\affiliation{%
		\institution{Purple Mountain Laboratories}
		\city{Nanjing}
		\postcode{211111}
		\country{China}}
\email{s.liu@seu.edu.cn}

\author{Weiliang Xie}
	\orcid{0009-0008-2542-8050}
\affiliation{%
	\institution{China Telecom Corporation Limited Technology Innovation Center}
	\city{Beijing}
	\country{China}}
\email{xiewl@chinatelecom.cn}

\author{Yongming Huang}
	\orcid{0000-0003-3616-4616}
\affiliation{%
 \institution{School of Information Science and Engineering, Southeast University}
 \city{Nanjing}
 \country{China}}
	\affiliation{%
		\institution{Purple Mountain Laboratories}
		\city{Nanjing}
		\postcode{211111}
		\country{China}}
\email{huangym@seu.edu.cn}

\begin{abstract}
Knowledge discovered from telecom data can facilitate proactive understanding of network dynamics and user behaviors, which in turn empowers service providers to optimize cellular traffic scheduling and resource allocation.
Nevertheless, the telecom industry still heavily relies on manual expert intervention. Existing studies have been focused on exhaustively explore the spatial-temporal correlations. However, they often overlook the underlying characteristics of cellular traffic, which are shaped by the sporadic and bursty nature of telecom services. Additionally, concept drift creates substantial obstacles to maintaining satisfactory accuracy in continuous cellular forecasting tasks. To resolve these problems, we put forward an online cellular traffic prediction method grounded in Multi-Grained Spatial-Temporal feature Complementarity (MGSTC). The proposed method is devised to achieve high-precision predictions in practical continuous forecasting scenarios. Concretely, MGSTC segments historical data into chunks and employs the coarse-grained temporal attention to offer a trend reference for the prediction horizon. Subsequently, fine-grained spatial attention is utilized to capture detailed correlations among network elements, which enables localized refinement of the established trend. The complementarity of these multi-grained spatial-temporal features facilitates the efficient transmission of valuable information. To accommodate continuous forecasting needs, we implement an online learning strategy that can detect concept drift in real-time and promptly switch to the appropriate parameter update stage. Experiments carried out on four real-world datasets demonstrate that MGSTC outperforms eleven state-of-the-art baselines consistently.
\end{abstract}




\received{20 February 2007}
\received[revised]{12 March 2009}
\received[accepted]{5 June 2009}

\maketitle

\section{Introduction}

The rapid proliferation of 5G networks and the ubiquity of mobile devices have ushered in an era of unprecedented connectivity, simultaneously intensifying the complexity of mobile data traffic management \cite{you2023toward}. This paradigm shift requires that network operators develop and implement robust methodologies for precise traffic prediction to ensure optimal network performance \cite{aouedi2025deep}. At its core, cellular traffic prediction involves forecasting traffic values given historical data, leveraging advanced statistical and machine learning techniques to discern patterns and trends \cite{cellular2022jiang}. Recent advancements in deep learning and artificial intelligence have opened new avenues for such predictive tasks.

The cellular traffic prediction approaches can be largely categorized into two groups: temporal and spatial-temporal. Temporal prediction focuses on individual network elements, such as single base stations, and utilizes their own historical data to understand localized dynamic patterns within specific network nodes.
Early studies are mostly built upon Recurrent Neural Network (RNN)-based models, exemplified by Long Short-Term Memory (LSTM) \cite{ka2023intelligent} and Gated Recurrent Unit (GRU) \cite{hu2022multi}. However, their inherent sequential properties hinder parallelization among training samples, which limits their scalability and efficiency. 
In contrast, Convolutional Neural Network (CNN)-based models allow for parallel extraction of temporal features \cite{ma2023cnn}, while Temporal Convolutional Networks (TCNs) further introduce causal convolutions and dilated structures, making them more effective for capturing long-range relationships \cite{bi2022tcn}.
Alternatively, benefited from self-attention mechanisms, Transformer-based architectures exhibit remarkable long-term modeling capabilities and improved interpretability \cite{zhou2021informer}.
Consequently, a series of temporal forecasting models based on Transformer backbone have emerged \cite{zhou2022fedformer, nie2023a}. Nevertheless, as these methods lack the explicit utilization of spatial correlations, they still have potential for further improvement.

The abundant information embedded in the spatial domain can also be leveraged to infer subsequent cellular traffic.
Such spatial-temporal prediction aims to forecast traffic across multiple network elements that exhibit spatial dependencies \cite{zhang2023crossformer}. This more complex approach takes into account the interconnected nature of mobile networks, recognizing that traffic patterns in one area can significantly influence those in adjacent or related regions. The spatial-temporal approach presents a more holistic view of network behavior, potentially offering more accurate predictions by capturing the intricate relationships between different network elements.
CNNs are initially employed to capture local spatial patterns for grid-structured data \cite{zang2021jointly, shen2021time-wise}. However, due to its inability to process non-Euclidean distances, Graph Neural Networks (GNNs) gradually become the mainstream \cite{zhao2022spatial}. As two representative methods, Graph Convolutional Networks (GCNs) \cite{zhu2022KST, shang2024spatial} employ fixed weights to aggregate neighboring information while Graph Attention Networks (GATs) \cite{huang2022learning, lin2021multivariate} achieve greater flexibility by adaptively updating weight coefficients through attention mechanisms. To reflect the time-varying spatial correlations in cellular networks, dynamic time warping is further employed to construct dynamic graphs as inputs of the subsequent GAT to learn timely spatial dependencies \cite{wang2023spatial}.
Recently, the diffusion convolutional GRU is also applied to spatial-temporal prediction tasks due to inherent compatibility with graph-structured data and achieves promising results \cite{xiao2025cellular}.

\begin{figure*}[t]
	\centering
	\includegraphics[width=0.95\linewidth]{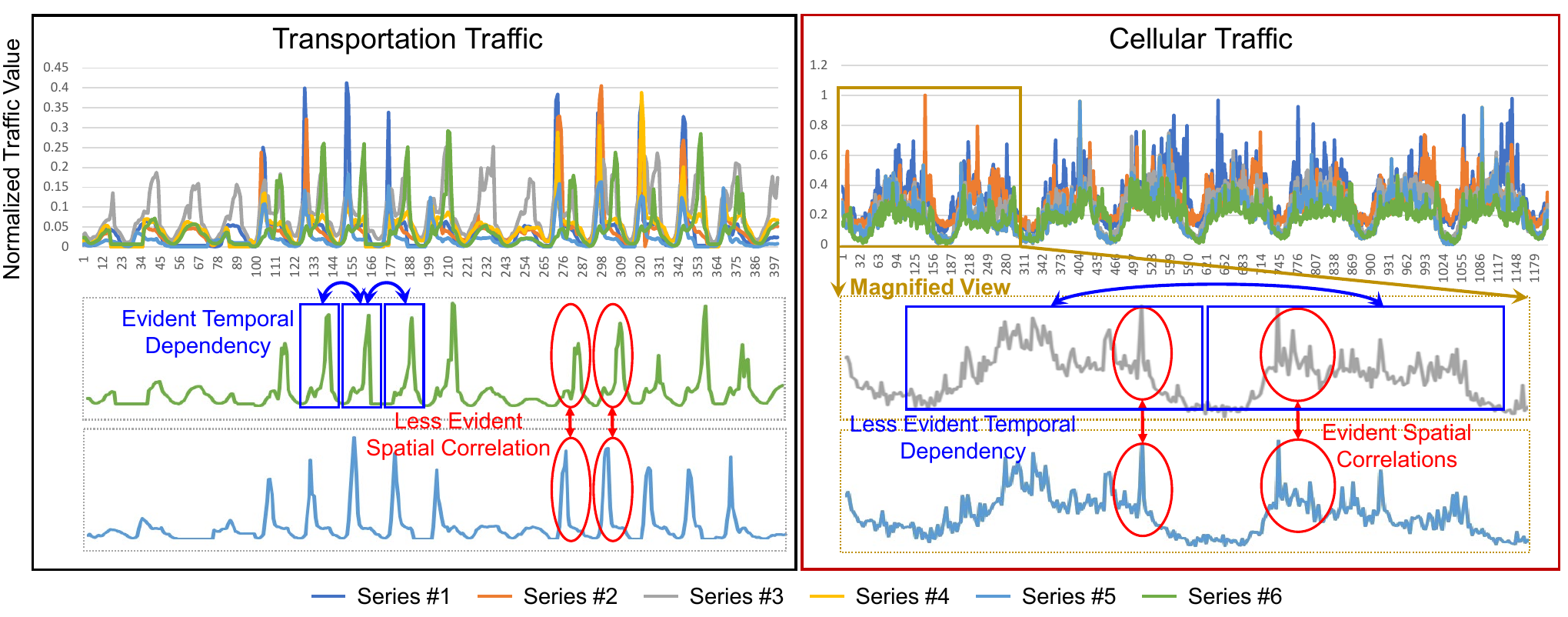}
	\caption{Left: Transportation traffic. Transportation flow exhibits evident temporal dependencies, with morning or evening rush hours consistently occurring at the same time each day. But traffic patterns of two neighboring intersections are less evident. Right: Cellular traffic with magnified views of two periods (two days). Cellular traffic only exhibits roughly similar trends over neighboring time intervals due to the sporadic and bursty nature of telecom services, whereas spatially adjacent series show a high degree of similarity in fluctuation details due to base station collaboration in high-demand areas.}
	\label{fig:intro}
\end{figure*}

However, most existing methods overly focus on exhaustive extraction of spatial-temporal dependencies in pursuit of precision improvement, but overlook the distinctive characteristics of cellular traffic. Fig.~\ref{fig:intro} illustrates a comparison of transportation and cellular traffic, with two series selected from each for detailed analysis.
We can observe that the temporal dynamic patterns of transportation flow are highly similar on a daily basis, with noticeable increases during the morning or evening rush hours. However, the spatial correlation is relatively weak, as indicated by the different peak positions in two series, where one intersection experience heavy traffic in the morning and the other in the evening rush hour.
In contrast, the magnified view of two periods (two days) highlights the extremely high dynamic nature of cellular traffic, as user demand peaks tend to surge within minutes and diminish swiftly.
Despite the daily periodicity, the occurrence of traffic peaks and troughs are rarely aligned with previous days, and intense fluctuations are more evidently propagated through spatial dimension.
This arises from the distinctive feature of telecom activities. When a Base Station (BS) becomes overloaded, neighboring BSs always provide support to alleviate the pressure, resulting in a concurrent traffic rise in adjacent regions.
Currently, predictive models tailored for cellular traffic considering its distinctive spatial-temporal characteristics are underdeveloped.

Another critical challenge is that offline trained models struggle to preserve satisfactory predictive performance in non-stationary cellular traffic streams.
The deployed models are expected to deliver continuous and accurate predictions. However, due to the underlying factors such as holiday schedule, crowd mobility and infrastructure modifications, cellular traffic streams tend to exhibit significant characteristic distribution changes over a period of time, which is termed as \textit{concept drift} \cite{zhou2023multi, zhang2024addressing}.
Although offline models excel on predetermined datasets, their fixed parameters restrict the scalability to practical streaming traffic data.
Several resolutions have been developed to address this issue. A straightforward method is conduct retraining using latest data at regular intervals, but this approach incurs intensive computational overhead and disregards previously learned knowledge \cite{miao2024unified}.
Alternatively, Experience Replay (ER) is a generic technique for online learning tasks, which enables the models to progressively accommodate new data while retaining knowledge from previous training \cite{chaudhry2019tiny, buzzega2020dark}. For instance, TrafficStream is a representative online prediction model for transportation flow that employs information replay and parameter smoothing to update model parameters annually \cite{chen2021trafficstream}.
Nevertheless, given the highly dynamic nature of cellular traffic, the occurrences of concept drift are both frequent and stochastic. Hence, the streaming cellular traffic needs to be continuously monitored in real time, with the prompt model updated accordingly.

Motivated by the above issue, we thoroughly investigate the distinct characteristic of cellular traffic. In light of the observed coarse-grained temporal dependencies and fine-grained spatial correlations, we propose an innovative online forecasting method based on Multi-Grained Spatial-Temporal feature Complementarity (MGSTC) tailored for cellular traffic streams. First, MGSTC segments historical sequences into large-scale chunks and applies Coarse-Grained Temporal Attention (CGTA) to capture chunk-wise temporal dependencies, which provide trend reference for prediction horizons. Second, the Fine-Grained Spatial Attention (FGSA) acts as a complement by extracting detailed spatial correlations, facilitating the localized refinement to the established trends.
Third, an online learning strategy is implemented to satisfy continuous prediction requirements, which is composed of a sensitive monitor for concept drift detection, and two corresponding parameter updating stages for adaptation.
The primary contribution of this paper can be summarized as:
\begin{itemize}
	\item We propose an innovative MGSTC method which segments the historical sequences into chunks and employs CGTA to provide trend reference for prediction window. Additionally, FGSA works in a complementary manner to propagate spatial correlations across network elements for localized refinement.
	\item MGSTC is further enhanced by an online learning strategy, which incorporates a sensitive monitor for concept drift detection and two corresponding parameter update stages for adaptation. Consequently, MGSTC can maintain the outstanding forecasting precision in the context of non-stationary cellular traffic streams.
	\item Extensive experiments are conducted on four distinctive real-world datasets, including Milan, Taiwan, AIIA and Bihar. Simulation results against state-of-the-art baselines demonstrate the superiority of MGSTC in terms of forecasting precision in both online and offline scenarios.
\end{itemize}

The rest of this paper is organized as follows. Section \uppercase\expandafter{\romannumeral2} provides a literature review from both offline and online cellular traffic forecasting methods. In Section \uppercase\expandafter{\romannumeral3}, we present a formal problem statement for online cellular traffic prediction and a thoroughly analysis of data characteristics. Then, the MGSTC is proposed in Section \uppercase\expandafter{\romannumeral4}, comprising the CGTA for capturing coarse-grained temporal dependencies, the FGSA for modeling fine-grained spatial correlations, and the online strategy designed for concept drift. Section \uppercase\expandafter{\romannumeral5} elaborates on the experimental results and demonstrates the exceptional performance of MGSTC. Finally, conclusion is drawn in Section \uppercase\expandafter{\romannumeral6}.

\section{Related Work}
\label{sec:rel_work}

Cellular traffic prediction serves as a key enabler of intelligent mobile network, such as providing prior information for load balancing among 5G BSs \cite{shang2024spatial} and drone deployment in wireless sensing networks \cite{betalo2024resource, betalo2024energy}.
The burgeoning progress in deep neural networks has revitalized cellular traffic prediction. Nowadays, forecasting methods can be classified into offline and online approaches, which will be elaborated on in the subsequent subsections.

\subsection{Offline Traffic Prediction Methods}

In the initial stage, LSTM is the most commonly adopted model for cellular forecasting \cite{he2020meta, wang2021data, ka2023intelligent}. Subsequently, TCNs are integrated with LSTM to capture multi-scale temporal relationships \cite{bi2022hybrid}.
An alternative approach is to incorporate spatial characteristics for performance improvement.
For instance, the authors of \cite{zhang2019deep} propose the STCNet, where a ConvLSTM is integrated with a CNN to capture spatial-temporal relationships and various external factors.
With the rising focus on non-Euclidean data, GNN-based methods have emerged as a prominent approach. In \cite{yu2021step}, a GRU works in conjunction with a GCN for fine-granular traffic prediction. An LSTM is also shown to be effectively coupled with a GAT to extract global spatio-temporal characteristics \cite{he2022graph}.
Moreover, by utilizing a graph-based TCN, user mobility pattern can also be exploited to enhance forecasting outcomes \cite{sun2022mobile}.

While the aforementioned RNN-based approaches have achieved remarkable advancement, they still struggle to model long-term dependencies and support parallelized training. Fortunately, Transformer overcomes these shortcomings through self-attention mechanism.
Such architectures have been widely applied to various spatial-temporal forecasting tasks, spanning transportation prediction \cite{kumar2024spatio, feng2022adaptive, zheng2024FGITrans}, pandemic forecasting \cite{ma2022hierarchical} and sequential recommender  \cite{xia2023multi}.
For cellular traffic prediction, a time-wise attention aided CNN is proposed in \cite{shen2021time-wise} to enhance long-term forecasting quality.
Additionally, self-attention mechanism has proven effective in both the temporal and spatial dimensions, leading to the development of STACN \cite{zhao2020spatial} and ST-Tran \cite{liu2021predictive}.
Subsequently, MVSTGN broadens the perspective of feature extraction by incorporating global spatial, global temporal and local spatial-temporal views \cite{MVSTGN2023Yao}.
To preserve shallow features during messaging process, a cascaded spatial-temporal Transformer GLSTTN is proposed with the collaboration of attention mechanisms and skip connections \cite{gu2023spatial}. Recently, authors of \cite{xiao2025cellular} introduce the DCG-MAM with multi-scale temporal modeling, but relies on a Gaussian kernel-based static graph, which limits its ability to capture dynamic spatial relationships. In the field of transportation flow prediction, several studies have explored multi-granularity approaches \cite{zhao2024stmgf, wei2024dynamic} . However, their overly fine-grained temporal modeling is not well-suited for highly dynamic cellular data, as it tends to amplify and propagate transient fluctuations.

\subsection{Online Traffic Prediction Methods}

Traditional prediction paradigm often runs in a batch learning fashion, which requires access to the entire training data prior to the learning task, and the training process is often conducted in an offline manner. These methods suffer from the fixed parameters, leading to limited scalability for cellular traffic streams. In contrast, online learning \cite{shui2023lifelong} is a method designed for sequentially arriving data, where the learner aims to continuously learn and update the best predictor for future data at each step.
At the nascent stage, ER is introduced to alleviate catastrophic forgetting and boost the model's generalization ability \cite{chaudhry2019tiny}. Subsequently, numerous variants of ER have been proposed, such as MIR \cite{aljundi2019online}, which selectively replays samples most prone to forgetting, and DER++ \cite{buzzega2020dark}, which incorporates the knowledge distillation techniques.
Utilizing the above ER technique, a renowned online prediction framework TrafficStream is developed for transportation flow, which performs annual updates of data and road topology with knowledge consolidation \cite{chen2021trafficstream}.
Alternatively, OneNet employs exponential gradient descent and reinforcement learning to adaptively balance the contributions of the cross-variable and cross-time branches in real time, enabling online prediction for data streams \cite{wen2023onenet}.
Despite their contributions, only a limited number of studies focus on the problem of online cellular traffic prediction. LNTP, as a representative approach, applies a simple back-propagation for each newly arrived network traffic data \cite{zhang2021LNTP} but neglects the concept drift challenge.

\section{Problem Formulation and Cellular Traffic Analysis}

\begin{figure*}[!t]
	\centering
	\includegraphics[width=0.9\linewidth]{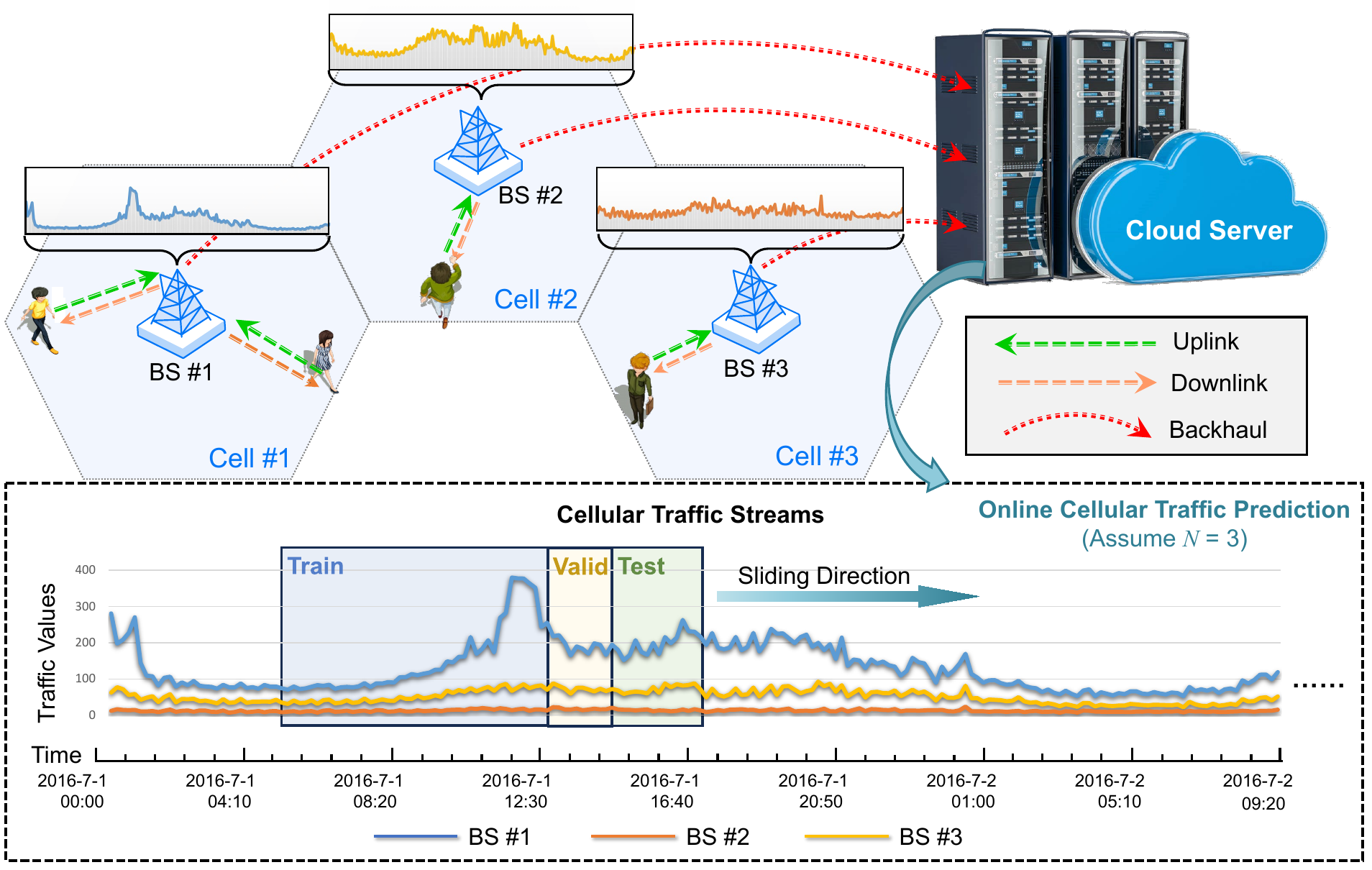}
	\caption{Typical mobile communication scenario and the online cellular prediction task.}
	\label{fig:scenario}
\end{figure*}

\subsection{Online Cellular Traffic Prediction}

A typical mobile communication scenario is depicted in Fig.~\ref{fig:scenario}. Each BS in the area keeps providing telecom services for mobile users, and records a series of cellular traffic data in real-time. Assuming that collected cellular traffic can be organized as a matrix $\mathbf{X} \in \mathbb{R}^{\Gamma \times N}$, where $\Gamma$ denotes time intervals and $N$ is the number of series. Additionally, cellular traffic vector observed at the $n$-th BS can be written as $\mathbf{x}_n \in \mathbb{R}^{\Gamma}$, while that collected at the $t$-th interval can be formulated as $\mathbf{x}^t \in \mathbb{R}^{N}$. Denote $x^t_n \in \mathbb{R}$ as the traffic value of the $n$-th series in the $t$-th interval, $\mathbf{x}_n$ can be formalized as $\mathbf{x}_n =[x_n^1, x_n^2, \ldots, x_n^\Gamma] $, and $\mathbf{x}^t=[x_1^t, x_2^t, \ldots, x_N^t]$.
These BSs send their traffic data to the cloud server via backhaul link at fixed intervals. The predictive model deployed in the cloud server continuously forecasts upcoming cellular traffic values based on historical data, which can be formally formulated as follows.

\begin{definition}
(\textbf{Online Cellular Traffic Prediction}).
\end{definition}

Considering a cellular traffic stream at time $t$, we retrieve historical data from the current moment $t$ going back $T$ steps: $\{\mathbf{x}^{t-T+1}, \mathbf{x}^{t-T+2},\ldots, \mathbf{x}^{t}\}$, which can be simplified as $\mathbf{X}^{(t-T+1:t)}$. We aim to forecast the future $\tau$ steps traffic value $\mathbf{Y}^{(t+1:t+\tau)}$ using the current mapping function $f^t(\cdot)$.  Namely,
\begin{equation}
	\hat{\mathbf{Y}}^{(t+1:t+\tau)} = f^t(\mathbf{X}^{(t-T+1:t)}),
\end{equation}
where $\hat{\mathbf{Y}}^{(t+1:t+\tau)}$ is the predictive value of $\mathbf{Y}^{(t+1:t+\tau)}$.
To specify the optimization direction for back-propagation, we choose the mean square error (MSE) loss to measure the discrepancy between the predicted results and the ground truth.
Then the loss function $\mathcal{L}\left(\mathbf{X}^{(t-T+1:t)}\right)$ can be calculated as
\begin{equation}
	\mathcal{L}\left(\mathbf{X}^{(t-T+1:t)}\right) = \mathbb{E} \left(\frac{1}{N} \sum^{N}_{n=1} \Vert \hat{\mathbf{Y}}^{(t+1:t+\tau)} -\mathbf{Y}^{(t+1:t+\tau)} \Vert_2^2\right),
\end{equation}
where $\Vert \cdot \Vert_2$ means the $l$-2 norm, and $\mathbb{E}(\cdot)$ denotes the expectation operation.
Subsequently, the mapping function $f^t(\cdot)$ updates its parameters according to the current loss $\mathcal{L}\left(\mathbf{X}^{(t-T+1:t)}\right)$, thus accommodating the evolving cellular characteristics and preserving satisfactory precision.
For notation clarity, we list the main notations and their interpretations in Table \ref{tab:notations}.
%Notably, in this scenario, the parameters of the prediction model $f^t(\cdot)$ keeps evolving as the testing samples progress until $t \leq \Gamma -\tau$.
\begin{table}
	\caption{Main Notations and Interpretations}
	\label{tab:notations}
	\begin{tabular}{cc}
		\toprule
		Notations & Interpretations \\
		\midrule
		$\mathbf{X}$ & Streaming cellular traffic data matrix \\
		$\Gamma$, $N$& Total time intervals and number of traffic series \\
		$x^t_n$ & The cellular traffic value of the $n$-th series at the $t$-th interval\\
		$\mathbf{x}^t$, $\mathbf{x}_n$ & Cellular traffic vector collected at the $t$-th interval and from the $n$-th series\\
		$T$, $\tau$ & The historical and prediction horizons\\
		$\mathbf{X}^{(t-T+1:t)}$ & Cellular traffic matrix from $t-T+1$ to $t$ \\
		$\mathbf{Y}^{(t+1:t+\tau)}$, $\hat{\mathbf{Y}}^{(t+1:t+\tau)}$ & The ground truth and predicted cellular traffic matrix from $t+1$ to $t+\tau$ \\
		$f^t(\cdot)$ & The online prediction model at time $t$ \\
		$C$, $S$, $M$ & The chunk length, stride length and the number of chunks\\
		$\mathbf{c}^m_n$, $\mathbf{C}_n$& The $m$-th chunk and the chunk matrix of the $n$-th series \\
		$\mathbf{G}$, $G$ & The aggregator matrix and the number of aggregator \\
		$D$ & The embedding dimension \\
		$\mathbf{H}$, $\mathbf{Z}$ & The temporal and spatial-temporal representation \\
		$\mathcal{B}$, $\mathcal{H}$ & The buffer and the historical repository \\
		$d$ & The concept drift threshold \\
		$\mathcal{L}^\dagger$, $\mathcal{L}^\ddagger$ & The fine-tuning loss and aggressive update loss \\
		$\eta^\dagger$, $\eta^\ddagger$ & The fine-tuning coefficient and aggressive update coefficient \\
		\bottomrule
	\end{tabular}
\end{table}

\subsection{Cellular Traffic Analysis}
Previous traffic prediction methods mostly favor exhaustive spatial-temporal feature extraction. These approaches repeatedly disseminate highly refined information across both dimensions, which
achieve marginally improvement at the expense of intensive calculation.
More concerningly, excessively fine-grained capture of temporal correlations can be counterproductive, as sporadic fluctuations in certain cellular traffic may mislead time-dependent extraction. Conversely, we argue that the granularity of temporal feature extraction can be relaxed, as spatial-temporal characteristic are mutually complementary.
\begin{definition}
	(\textbf{Multi-Grained Spatial-Temporal Complementarity}).
Let $\mathcal{F}_\mathrm{S}^{g_1}$ denote the spatial feature space extracted under granularity $g_1$, and $\mathcal{F}_\mathrm{T}^{g_2}$ denote the temporal feature space under granularity $g_2$. Given that spatial and temporal features exhibit partial redundancy:
\begin{equation}
	\mathcal{F}_\mathrm{S}^{g_1} \cap \mathcal{F}_\mathrm{T}^{g_2} \neq \varnothing.
	\end{equation}
	We define multi-grained spatial-temporal complementarity as the property that for a relaxed granularity $g'_1 > g_1$ or $g'_2 > g_2$, the spatial and temporal feature space $\mathcal{F}_\mathrm{S}^{g'_1}$ and $\mathcal{F}_\mathrm{T}^{g'_2}$ still satisfies:
	\begin{equation}
		\mathcal{F}_\mathrm{S}^{g'_1} \cap \mathcal{F}_\mathrm{T}^{g'_2} \approx \mathcal{F}_\mathrm{Full},
	\end{equation}
	where $\mathcal{F}_\mathrm{Full}$ represents the complete feature space. This implies that even with coarser temporal features, meaningful complementarity to spatial features is still retained.
\end{definition}
\begin{figure*}[!t]
	\centering \subfigure[]{\includegraphics[width=0.44\linewidth]{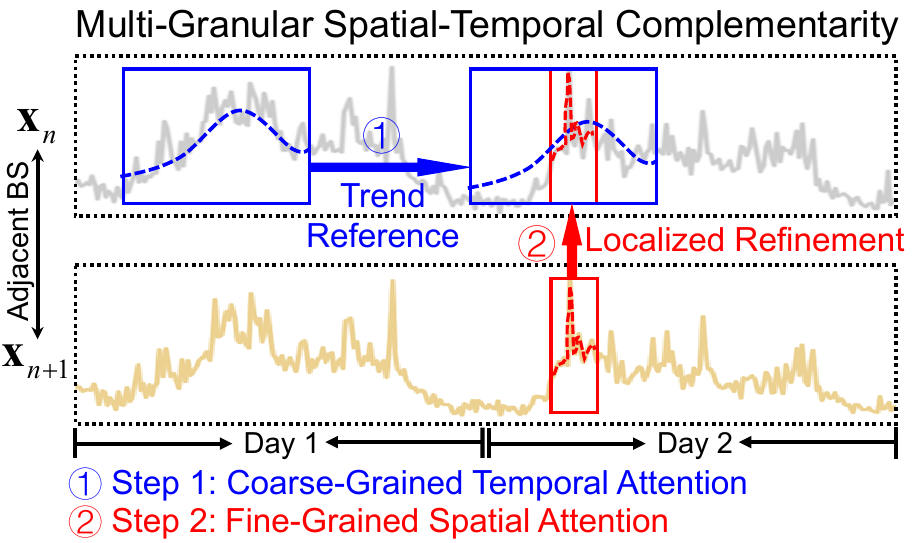}\label{fig:coarse_fine}}
	\subfigure[]{\includegraphics[width=0.55\linewidth, trim=0 -40 0 0,clip]{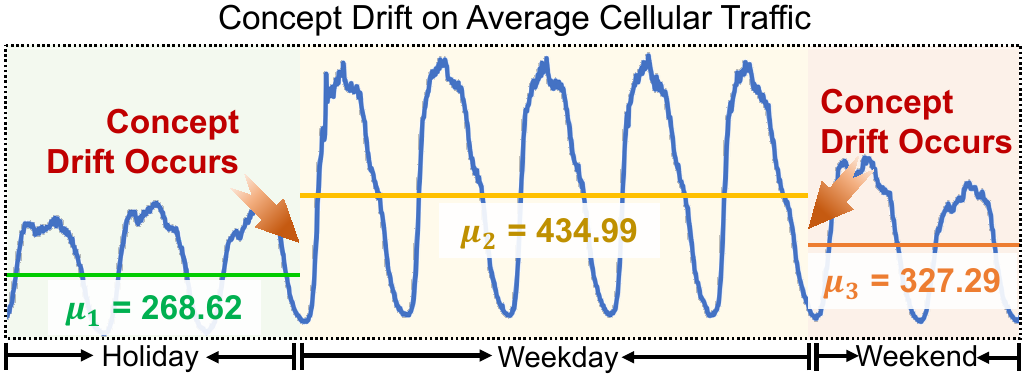}\label{fig:concept_drift}}
	\vspace{-1em}
	\caption{Cellular Traffic Analysis. (a) Multi-grained spatial-temporal complementarity attribute. The cellular traffic from two adjacent BSs denoted by $\mathbf{x}_n$ and $\mathbf{x}_{n+1}$ are selected for visualization. Given that the traffic of Day 1 and Day 2 exhibits similar trends, coarse-grained attention is applied along the time dimension to provide trend reference. Based on the similar burst fluctuations of $\mathbf{x}_n$ and $\mathbf{x}_{n+1}$, fine-grained spatial attention is employed for localized refinement. (b) Illustration of cellular concept drift. We display the average cellular traffic $\frac{1}{N}\sum^{N}_{n=1}\mathbf{x}_{n}$ of Milan dataset (a popular telecom dataset), which shows the occurrence of concept drift when the mean (as one of the statistical characteristics) undergoes noticeable changes, where $\mu_1=268.62$ for holidays, $\mu_2 =434.99$ for weekdays and $\mu_3=327.29$ for weekends.}
	\label{fig:analysis}
\end{figure*}
Building upon the analysis, we aim to capture coarse-grained temporal dependencies to provide trend reference for prediction horizon, while extract fine-grained spatial correlations for localized refinement as illustrated in Fig.~\ref{fig:coarse_fine}. The complementarity of the multi-grained perspective for cellular feature extraction enables the effective propagation of valuable information under reduced computational overhead, which is potential to achieve improved predictions. On the other hand, the network underlying factors are subject to ongoing evolution over time, such as holiday schedule, crowd mobility, customer preferences shifts, and infrastructure adjustments, which lead to the cellular concept drift.
\begin{definition}
	(\textbf{Cellular Concept Drift}).
	Cellular concept drift indicates unpredictable changes of statistical characteristics, or volatile evolution of spatial-temporal relationships over time in cellular traffic streams. Namely, concept drift is considered to have occurred between $t_1$ and $t_2$ if Eq. \eqref{eq:concept_drift} is satisfied.
	\begin{equation}
		\exists t_1, t_2 \in \Gamma \quad \text{such that}\quad P(\mathbf{X}^{(t_1-T+1:t_1)}, \mathbf{Y}^{(t_1+1:t_1+\tau)}) \neq P(\mathbf{X}^{(t_2-T+1:t_2)}, \mathbf{Y}^{(t_2+1:t_2+\tau)}),
		\label{eq:concept_drift}
	\end{equation}
	where $P(\mathbf{X}^{(t-T+1:t)}, \mathbf{Y}^{(t+1:t+\tau)})$ represents the joint probability distribution of $\mathbf{X}^{(t-T+1:t)}$ and $\mathbf{Y}^{(t+1:t+\tau)}$.
\end{definition}

We depict an obvious cellular concept drift in Fig. \ref{fig:concept_drift}, which is related to the holiday schedule. The significant transitions of average traffic value during holiday ($\mu_1$), weekday ($\mu_2$) and weekend ($\mu_3$) indicate the occurrence of concept drift.
In fact, cellular concept drift is particularly prevalent due to the highly dynamic and deeply coupled nature of mobile networks. Hence, predictive models necessitate real-time updates to address this issue, in order to prevent the progressive accumulation of errors.

\section{Methodology}

\begin{figure*}[t]
	\centering
	\includegraphics[width=0.96\linewidth]{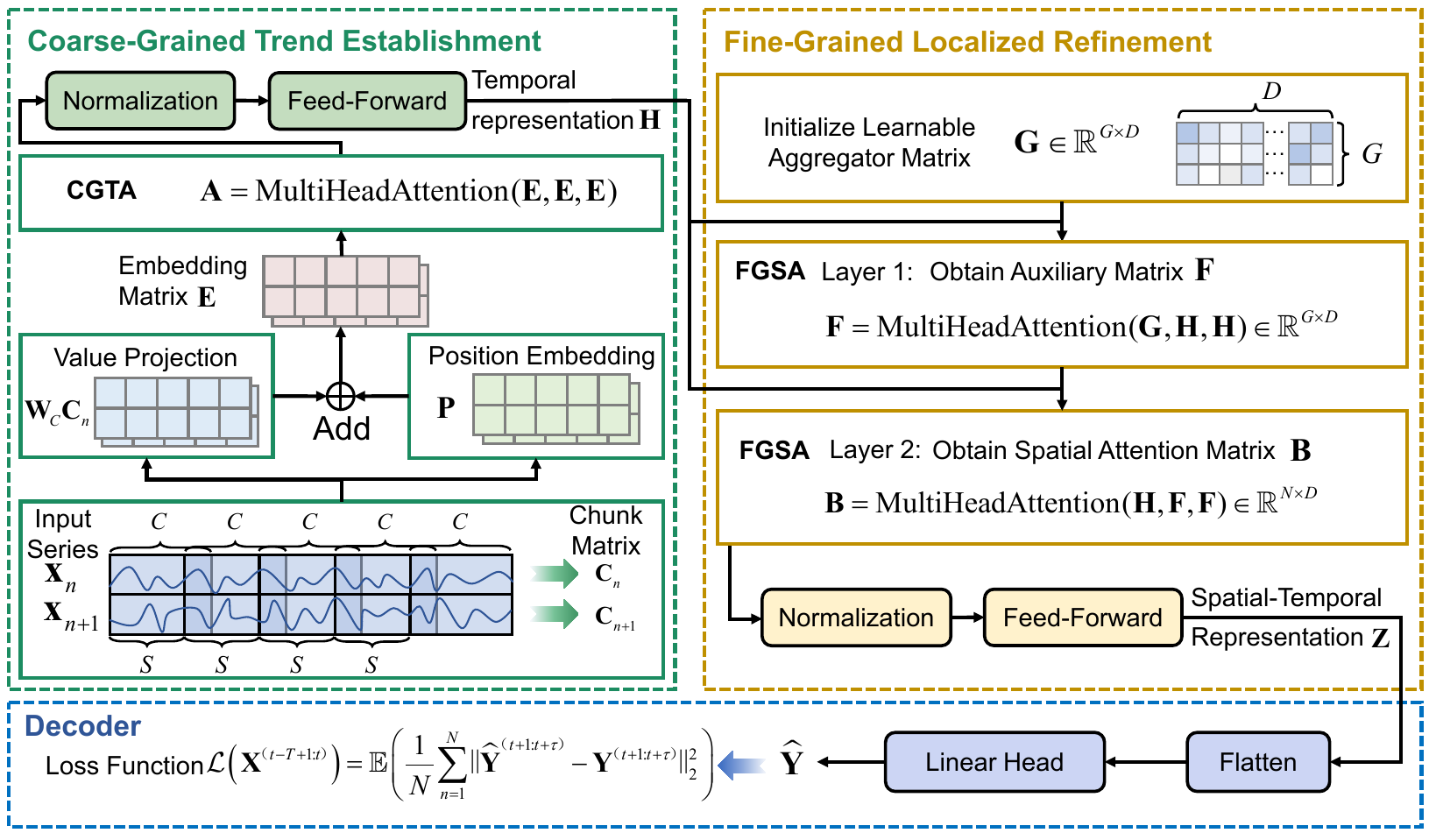}
	\caption{Network architecture of MGSTC.}
	\label{fig:offline_model}
\end{figure*}

In this section, we elaborate on three components of MGSTC, including coarse-grained trend establishment, fine-grained localized refinement, and online learning strategy. First, the input long series are segmented into chunks and then fed into a CGTA, which provide trend reference for subsequent prediction horizons. Second, the detailed spatial information are complemented with temporal features through a FGSA, facilitating the localized refinement on constructed trends. Finally, we further implement an online learning strategy to detect and adapt to the cellular concept drift, thus preserving exceptional accuracy in continuous forecasting.
The network architecture of MGSTC is illustrated in Fig. \ref{fig:offline_model}.

\subsection{Coarse-grained Trend Establishment}
Apparent daily periodicity of cellular traffic can be leveraged to provide trend reference for prediction horizon. Hence, we employ a CGTA to capture low-granularity temporal dependencies. 
In order to preserve the information at the tail of each series, we adopt a padding technique by repeating the last value $S$ times.
The input series are then segmented into multiple overlapped or non-overlapped chunks. Specifically, for the $n$-th cellular traffic series $\mathbf{x}_n$, the $m$-th chunk $\mathbf{c}^m_n$ can be obtained as
\begin{equation}
	\mathbf{c}^m_n = \Vert_{t=(m-1)S+1}^{(m-1)S+C} x^t_n \in \mathbb{R}^C,
\end{equation}
where $\Vert$ means the concatenation operation, $C$ and $S$ denote the chunk length and the non-overlapping region between two consecutive chunks, i.e., stride length.
The segmentation process convert a $T$-length series $\mathbf{x}_n$ to a chunk matrix $\mathbf{C}_n$ consisting of $M$ chunks $\mathbf{C}_n = [\mathbf{c}^1_n, \mathbf{c}^2_n, \ldots, \mathbf{c}^M_n] \in \mathbb{R}^{M\times C}$, and $M$ can be calculated as
\begin{equation}
	M=\Bigg \lfloor \frac{T-C}{S} \Bigg \rfloor + 2.
\end{equation}
Different from the other segment methods, such as patching technique, we adopt a much longer segmentation length to achieve the coarse-grained feature extraction, i.e., $C$ is relatively large. This arises from the fact that larger segments can better eliminate drastic fluctuations along the time dimension and convey more robust trend information.
Then the obtained chunk matrix $\mathbf{C}_n$ is input into the value projection as $\mathbf{W}_C \mathbf{C}_n$, where $\mathbf{W}_C$ is the learnable projection matrix. Denote the embedding dimension as $D$ and the position of chunks as $\mathrm{pos}$, the position embedding matrix $\mathbf{P}$ can be derived by the trigonometric function, whose even position is calculated as $	\mathbf{P}_{ 2i}=\sin\left({\frac{\mathrm{pos}}{10000^{2i/D}}}\right)$, while the odd position is $ \mathbf{P}_{ 2i+1}=\cos\left({\frac{\mathrm{pos}}{10000^{2i/D}}}\right)$.
The output representation $\mathbf{E}$ of the embedding layer can be written as the sum of both
\begin{equation}
	\mathbf{E} = \mathbf{W}_C \mathbf{C}_n + \mathbf{P} \in \mathbb{R}^{M \times D}.
\end{equation}

Then the CGTA is applied to the embedding matrix $\mathbf{E}$. Denote the learnable parameter matrices for query, key and value matrices as $\mathbf{W}_Q$, $\mathbf{W}_K$ and $\mathbf{W}_V$, the query matrix $\mathbf{Q}$, key matrix $\mathbf{K}$ and value matrix $\mathbf{V}$ can be calculated as $\mathbf{Q} =\mathbf{E} \mathbf{W}_Q$, $\mathbf{K} =\mathbf{E} \mathbf{W}_K$ and $\mathbf{V} =\mathbf{E} \mathbf{W}_V$. Then, the CGTA is obtained by applying the multi-head self-attention on $\mathbf{E}$, which can be expressed as
\begin{equation}
\mathbf{A} = \operatorname{MultiHeadAttention}(\mathbf{E}, \mathbf{E}, \mathbf{E}) = \Vert_{h=1}^H \left( \operatorname{Softmax}\left(\frac{\mathbf{Q}\mathbf{K}^\mathsf{T}}{\sqrt{d_k}}\right) \mathbf{V} \right),
\end{equation}
where $\operatorname{MultiHeadAttention}(\cdot)$ represents the multi-head self-attention operation, $d_k$ indicates the scale factor and $H$ is the number of attention head. The results of self-attention are concatenated $H$ times to obtain the temporal attention matrix $\mathbf{A}$.
Subsequently, $\mathbf{A}$ is input to a normalization layer with a residual connection. Suppose that $\mathbf{A}$ contains $I$ parameters, the mean $\mu_A$ and variance $\sigma^2_A$ of $(\mathbf{A}+\mathbf{E})$ are first calculate as
\begin{equation}
	\mu_A = \frac{1}{I}\sum^{I}_{i=1}(\mathbf{A}_i+\mathbf{E}_i), \quad \sigma^2_A = \frac{1}{I}\sum^{I}_{i=1}\left((\mathbf{A}_i+\mathbf{E}_i)-\mu_A \right)^2,
\end{equation}
where $\mathbf{A}_i$ and $\mathbf{E}_i$ are the $i$-the parameter of $\mathbf{A}$ and $\mathbf{E}$. The value of $(\mathbf{A}+\mathbf{E})$ can be standardized as
\begin{equation}
	\tilde{\mathbf{H}} = \gamma \frac{(\mathbf{A}+\mathbf{E})-\mu_A}{\sqrt{\sigma^2_A+\epsilon}} + \beta,
\end{equation}
where $\epsilon$ is a small positive number to prevent division by zero, $\gamma$ and $\beta$ are two learnable parameters. Then $\tilde{\mathbf{H}}$ is fed into a feed-forward layer with a residual connection as
\begin{equation}
	\mathbf{H} = \sigma (\mathbf{W}_H\cdot \tilde{\mathbf{H}} + \mathbf{b}_H) + \tilde{\mathbf{H}},
\end{equation}
where $\sigma$ is the activation function, $\mathbf{W}_H$ and $\mathbf{b}_H$ are the weight coefficient and bias vector. The output $\mathbf{H}$ is the final temporal representation.

\subsection{Fine-grained Localized Refinement}

Upon the determination of general trend by CGTA, localized refinement can be performed based on the FGSA across neighboring series as a complement.
To achieve more meticulous spatial correlation, the FGSA is composed of two cascaded multi-head self-attention layers. 
Unlike the conventional two-layer stacked full spatial attention, we introduce a learnable aggregator $\mathbf{G}\in\mathbb{R}^{G\times D}$, which serves as the query matrix in the first layer. 
With the incorporation of $\mathbf{G}$, FGSA not only maintains the attention depth to capture profound relationships, but also significantly reduces its complexity from the full spatial attention's $\mathcal{O}(N^2D)$ to $\mathcal{O}(NGD)$. Configuring $G\ll N$ allows the quadratic complexity to be reduced to linear, especially in scenarios involving numerous sequences.
The auxiliary matrix $\mathbf{F}$ generated by the first attention layer of FGSA can be derived as
\begin{equation}
	\mathbf{F} =\operatorname{MultiHeadAttention}(\mathbf{G}, \mathbf{H}, \mathbf{H})=\Vert_{h=1}^H \left( \operatorname{Softmax}\left(\frac{(\mathbf{G}\mathbf{W}'_Q)(\mathbf{H}\mathbf{W}'_K)^\mathsf{T}}{\sqrt{d_k}}\right) (\mathbf{H}\mathbf{W}'_V) \right) \in \mathbb{R}^{G\times D},
\end{equation}
where $\mathbf{W}'_Q$, $\mathbf{W}'_K$ and $\mathbf{W}'_V$ are the first layer's learnable parameter matrices. $\mathbf{G}$ serves as the query matrix while $\mathbf{H}$ is used for key and value matrices. Subsequently, the second attention layer of FGSA is applied to $\mathbf{H}$ and $\mathbf{F}$ as
\begin{equation}
	\mathbf{B} =\operatorname{MultiHeadAttention}(\mathbf{H}, \mathbf{F}, \mathbf{F}) = \Vert_{h=1}^H \left( \operatorname{Softmax}\left(\frac{(\mathbf{H}\mathbf{W}''_Q)(\mathbf{F}\mathbf{W}''_K)^\mathsf{T}}{\sqrt{d_k}}\right) (\mathbf{F}\mathbf{W}''_V) \right) \in \mathbb{R}^{N\times D},
\end{equation}
where $\mathbf{W}''_Q$, $\mathbf{W}''_K$ and $\mathbf{W}''_V$ are the second learnable parameter matrices. In this attention layer, $\mathbf{H}$ functions as the query matrix while $\mathbf{F}$ is regarded as the key and value matrices. $\mathbf{B}$ is the final output of FGSA.
Similar to the temporal module, the output $\mathbf{B}$ also passes a normalization layer with residual connection to produce $\tilde{\mathbf{Z}}$, and the final spatial-temporal representation matrix $\mathbf{Z}$ can be derived through a feed-forward layer with a residual connection as $\mathbf{Z} =  \sigma(\mathbf{W}_Z \cdot \tilde{\mathbf{Z}} + \mathbf{b}_Z) + \tilde{\mathbf{Z}}$.

The decoder used in our model is just a flatten layer $\operatorname{Flatten(\cdot)}$ followed by a linear head, with $\mathbf{W}$ denotes the weight coefficient and $\mathbf{b}$ represents the bias vector, the derivation within the decoder can be written as
\begin{equation}
	\hat{\mathbf{Y}}^{(t+1:t+\tau)} = \mathbf{W}\cdot \operatorname{Flatten}\left(\mathbf{Z}\right) + \mathbf{b}.
\end{equation}

\subsection{Online Learning Strategy}

\begin{figure*}[t]
	\centering
	\includegraphics[width=0.8\linewidth]{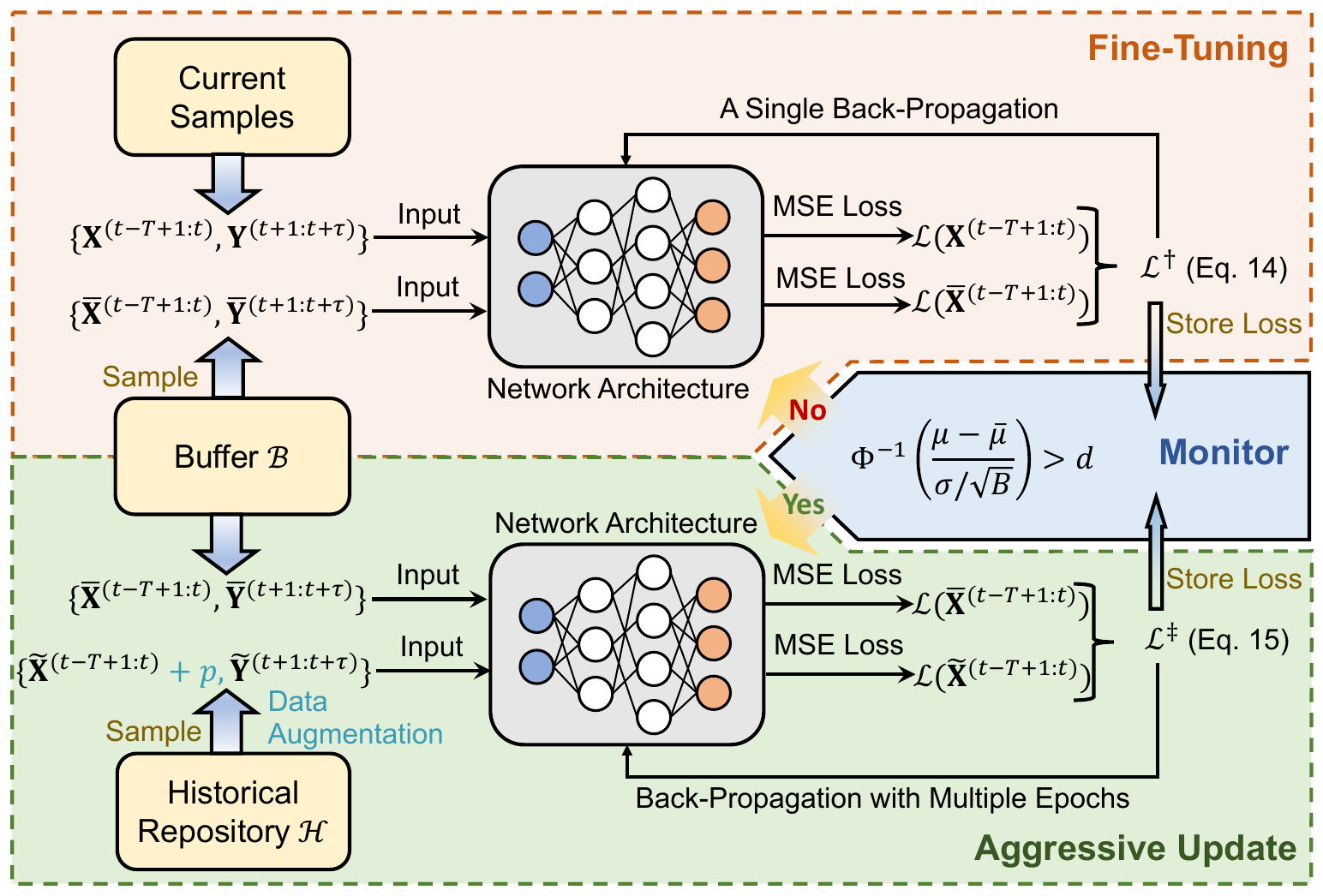}
	\caption{Overall pipeline of the MGSTC's online learning strategy.}
	\label{fig:online_strategy}
\end{figure*}

To accommodate the continuous forecasting requirement, we devise an online learning strategy to augment MGSTC with the detection and adaptation capabilities in response to cellular concept drift.
The overall pipeline of the online learning strategy is shown in Fig. \ref{fig:online_strategy}, which encompasses a monitor together with two parameter update stages, namely fine-tuning and aggressive update.
The monitor is designed to identify the occurrence of concept drift. If no drift is detected, MGSTC adopts fine-tuning for gradual parameter updates. Once concept drift occurs, MGSTC switches to the aggressive update phase immediately.
Next, we will provide a detailed elaboration for these three components and the complete training and inference process is summarized in Algorithm~\ref{alg1}.

\subsubsection{Concept Drift Monitor}

In the case of cellular traffic samples, the training loss of stationary data always follows a normal distribution $\mathcal{N}(\mu, \sigma)$, while the loss of non-stationary data will gradually deviate from this distribution during testing process.
Hence, the occurrence of cellular concept drift can be determined by performing hypothesis testing on the loss values.

Specifically, we establish a buffer $\mathcal{B}$ to store the latest $B$ test losses $\{\mathcal{L}_b\}_{b=1}^B$. The mean value of losses in $\mathcal{B}$ can be calculated as $\bar{\mu}=\sum^B_{b=1}\mathcal{L}_b$.
For the current test traffic $\mathbf{X}^{(t-T+1:t)}$, the mean and variance values are also calculated as $\mu$ and $\sigma$.
Given a pre-defined drift threshold $d$, the criteria for concept drift is defined as
\begin{equation}
	D(\mathbf{X}^{(t-T+1:t)}) = \left\{
		\begin{array}{ll}
			1 \quad \text{if}\quad \Phi^{-1}\left (\frac{\mu-\bar{\mu}}{\sigma/\sqrt{B}}\right) > d, \\
		0 \quad \text{otherwise},
		\end{array}
		\right.
	\label{eq:criteria}
\end{equation}
where $D(\cdot)$ is the drift indicator, $\Phi^{-1}(\cdot)$ is the inverse function of standard Gaussian cumulative distribution function $\Phi(\cdot) = \int^{\frac{\mu-\bar{\mu}}{\sigma/\sqrt{B}}}_{-\infty}e^{-t^2/2}/\sqrt{2\pi}dt$. $D(\mathbf{X}^{(t-T+1:t)})=0$ indicates no drift has been identified, allowing MGSTC to proceed with updates by fine-tuning. Conversely, $D(\mathbf{X}^{(t-T+1:t)})=1$ implies the detection of drift, and MGSTC is taken over by aggressive update.

\begin{algorithm}[H]
	\renewcommand{\algorithmicrequire}{\textbf{Input:}}
	\renewcommand{\algorithmicensure}{\textbf{Output:}}
	\caption{Training and Inference Process of MGSTC}
	\label{alg1}
	\begin{algorithmic}[1]
		\STATE Initialize a FIFO buffer $\mathcal{B}$ with a capacity limit of 100 samples, and a FIFO historical repository $\mathcal{H}$ with a capacity limit of 256 samples.
		\STATE Train the model using the training set; \COMMENT{Online inference starts upon completion of training}
		\FOR {Each batch of test samples} 
		\STATE Feed $\mathbf{X}^{(t-T+1:t)}$ into the model and obtain $\mathcal{L}(\mathbf{X}^{(t-T+1:t)})$;
		\IF {$\mathcal{B} = \emptyset$} 
		\STATE $\mathcal{L}(\bar{\mathbf{X}}^{(t-T+1:t)}) \leftarrow 0$; 
		\ELSE 
		\STATE Get samples from $\mathcal{B}$ and feed them into the model to obtain $\mathcal{L}(\bar{\mathbf{X}}^{(t-T+1:t)})$;
		\ENDIF 
		\STATE $\mathcal{L}^\dagger = \mathcal{L}(\mathbf{X}^{(t-T+1:t)}) + \eta^\dagger \mathcal{L}(\bar{\mathbf{X}}^{(t-T+1:t)})$; \COMMENT{Corresponding to Eq. \eqref{eq:fine_tuning}}
		\STATE Add samples $\mathbf{X}^{(t-T+1:t)}$ to $\mathcal{B}$;
		\IF{Concept drift occurs $D(\mathbf{X}^{(t-T+1:t)})=1$} 
		\STATE Reset the monitor;
		\STATE Get samples from $\mathcal{B}$ and feed them into the model to obtain $\mathcal{L}(\bar{\mathbf{X}}^{(t-T+1:t)})$;
		\IF{$\mathcal{H}=\emptyset$}
		\STATE $\mathcal{L}(\tilde{\mathbf{X}}^{(t-T+1:t)}) \leftarrow 0$;
		\ELSE
		\STATE Get samples from $\mathcal{H}$ and feed them into the model to obtain $\mathcal{L}(\tilde{\mathbf{X}}^{(t-T+1:t)})$;
		\ENDIF
		\STATE $\mathcal{L}^\ddagger = \mathcal{L}(\bar{\mathbf{X}}^{(t-T+1:t)}) + \eta^\ddagger \mathcal{L}(\tilde{\mathbf{X}}^{(t-T+1:t)})$; \COMMENT{Corresponding to Eq. \eqref{eq:aggressive_update}}
		\STATE Add all samples stored in $\mathcal{B}$ to $\mathcal{H}$ adhering to the FIFO principle;
		\STATE Reset the buffer $\mathcal{B}$.
		\ENDIF
		\ENDFOR
	\end{algorithmic}  
\end{algorithm}

\subsubsection{Fine-Tuning Stage}

In the fine-tuning stage, MGSTC's parameters are moderately adjusted based on the current cellular traffic sample $\{\mathbf{X}^{(t-T+1:t)}, \mathbf{Y}^{(t+1:t+\tau)}\}$ and the randomly selected sample $\{\bar{\mathbf{X}}^{(t-T+1:t)}, \bar{\mathbf{Y}}^{(t+1:t+\tau)}\}$ from buffer $\mathcal{B}$. Then these samples are fed into the prediction model, generating two losses $\mathcal{L}(\mathbf{X}^{(t-T+1:t)})$ and $\mathcal{L}(\bar{\mathbf{X}}^{(t-T+1:t)})$, respectively.
The weighted sum of these two losses constitutes the total loss function $\mathcal{L}^\dagger$ in the fine-tuning stage,
\begin{equation}
	\mathcal{L}^\dagger = \mathcal{L}(\mathbf{X}^{(t-T+1:t)}) + \eta^\dagger \mathcal{L}(\bar{\mathbf{X}}^{(t-T+1:t)}),
	\label{eq:fine_tuning}
\end{equation}
where $\eta^\dagger$ is the fine-tuning coefficient. A single back-propagation is performed based on loss $\mathcal{L}^\dagger$ to slightly adjust the MGSTC's parameters.

\subsubsection{Aggressive Update Stage}

Once a concept drift alert is triggered, i.e., $D(\mathbf{X}^{(t-T+1:t)})=1$, MGSTC transitions from fine-tuning to the aggressive update immediately, which conduct drastic adjustments to model parameters.
In this stage, MGSTC is trained using augmented historical data to enhance its generalization ability. Hence, a historical repository $\mathcal{H}$ needs to be established. Different from $\mathcal{B}$, $\mathcal{H}$ contains more earlier samples with a larger storage capacity, i.e., $|\mathcal{H}| \gg |\mathcal{B}|$.
Next, a portion of cellular samples $\{\tilde{\mathbf{X}}^{(t-T+1:t)}, \tilde{\mathbf{Y}}^{(t+1:t+\tau)}\}$ are selected from $\mathcal{H}$, and then added by a perturbation $p$ to produce the augmented historical samples $\{\tilde{\mathbf{X}}^{(t-T+1:t)} + p, \tilde{\mathbf{Y}}^{(t+1:t+\tau)}\}$.
The proof of the effectiveness of such data augmentation is provided in the Appendix.
In this stage, samples $\{\bar{\mathbf{X}}^{(t-T+1:t)}, \bar{\mathbf{Y}}^{(t+1:t+\tau)}\}$ and $\{\tilde{\mathbf{X}}^{(t-T+1:t)}, \tilde{\mathbf{Y}}^{(t+1:t+\tau)}\}$ are simultaneously fed into the prediction model, generating losses $\mathcal{L}(\bar{\mathbf{X}}^{(t-T+1:t)})$ and $\mathcal{L}(\tilde{\mathbf{X}}^{(t-T+1:t)})$, respectively. The weighted sum loss $\mathcal{L}^\ddagger$ can be calculated with the introduction of the aggressive coefficient $\eta^\ddagger$,
\begin{equation}
	\mathcal{L}^\ddagger = \mathcal{L}(\bar{\mathbf{X}}^{(t-T+1:t)}) + \eta^\ddagger\mathcal{L}(\tilde{\mathbf{X}}^{(t-T+1:t)}).
	\label{eq:aggressive_update}
\end{equation}
Notably, different from the fine-tuning stage, this stage involves multiple training epochs to guarantee adequate changes in MGSTC's parameters.

\section{Experiments}

In this section, extensive experimental results are presented to evaluate the effectiveness of MGSTC. We first introduce the datasets and experimental settings. Then, forecasting precision and efficiency comparisons against state-of-the-art models are provided for both offline and online tasks. Subsequently, several key parameters are adjusted to illustrate their impacts, and the ablation experiments are conducted to demonstrate the necessity of each component.

\subsection{Data Description}

Four popular telecom datasets are included for evaluation, i.e., Milan, Taiwan,  AIIA and Bihar. Their statistics are summarized in Table~\ref{tab:datasets}.
\begin{table}[t]
	\caption{Statistics of four real-world public datasets.}
	\label{tab:datasets}
	\begin{tabular}{c|cccc}
		\toprule
		Datasets & Number of Series & Samples & Timespan & Timeslot \\
		\midrule
		Milan & 900 & 8928 & 2013-11-01 00:00 - 2014-01-01 23:50 & 10 min\\
		\midrule
		Taiwan & 6 & 50000 & 2020-01-01 00:05 - 2020-06-22 14:40 & 5 min \\
		\midrule
		AIIA & 3 & 16000 & 2017-01-01 00:00 - 2018-10-29 15:00 & 1 h \\
		\midrule
		Bihar & 25 & 3501 & 2023-05-05 12:45 - 2023-06-10 23:45 & 15 min \\
		\bottomrule
	\end{tabular}
\end{table}

\begin{figure}[t]
	\centering
	\includegraphics[width=0.65\linewidth]{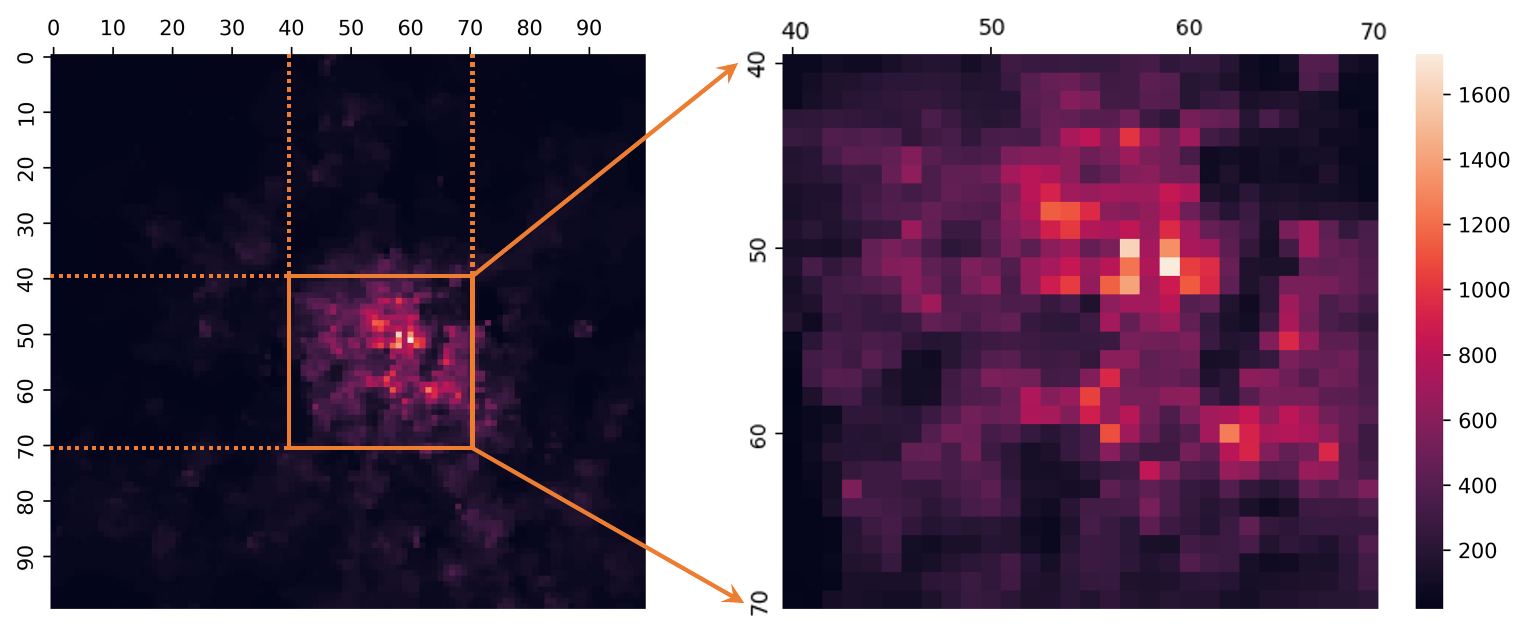}
	\caption{Heat map of average total cellular traffic in Milan. Complete $100 \times 100$ grids (left) and the selected $30 \times 30$ center cells (right).}
	\label{Milan10000-900}
\end{figure}

\textbf{Milan:}
This publicly available dataset provided by Telecom Italia \cite{barlacchi2015multi} has become the most popular public datasets for validation on mobile communication tasks, such as cellular traffic prediction, distributed caching, etc.
This dataset records the cellular traffic volume of Milan at 10-minute intervals, where geographical area is divided into $100 \times 100$ cells with each covering an area of $235\;{\rm{m}} \times 235\;{\rm{m}}$.
Each record contains five categories of telecom activities, which are SMS-in, SMS-out, Call-in, Call-out and Internet traffic activity. In this experiment, these five activities are aggregated into a total traffic volume.
The complete dataset contains two months data from 01/11/2013 to 01/01/2014 (62 days) with 8928 entries. The first five days data are used for training, the following two days for validation, and the remaining for testing, i.e., the ratio of training, validation, and testing is 5:2:55.
Fig.~\ref{Milan10000-900} illustrates the heat map of the average total traffic volume in Milan. It indicates that the cellular traffic maintains a substantial volume in the center regions. Considering the concentrations on high-demand areas, we select the center regions ($30 \times 30$ grids) for subsequent evaluation.

\textbf{Taiwan:} This dataset, first presented in \cite{lin2021multivariate}, records the cellular traffic at six road intersections (e.g., around the train station and college).
The quantity of outdoor terminals located at these intersections is aggregated in the unit time step (5 min). In this dataset, a total of 52415 entries was collected from 01/01/2020 00:05 to 30/06/2020 23:55. We select the first 50000 entries out of them, and divided the train/validation/test sets in a ratio of 3:2:45.

\textbf{AIIA:} This dataset comes from the ``AIIA Home Network Competition: Network Traffic Forecasting'', which records the hourly granularity data of three different regions A, B and C from 00:00 on January 1, 2017 to 23:00 on November 15, 2018 \cite{da2021dynamic}. Similarly, the first 16000 entries are selected among a total of 16416 entries with the division ratio as 4:2:10.

\textbf{Bihar:} This dataset collected in Bihar, India during May and June of 2023 comes from the Kaggle \footnote{ https://www.kaggle.com/datasets/suraj520/cellular-network-analysis-dataset}, which provide realistic signal throughput for 3G, 4G, 5G, and LTE cellular networks. We partition the entire collection area into 25 cells according to their latitude and longitude coordinates, and aggregate the throughput of various networks into 15-minute intervals.
Given the sparse and highly random short-duration traffic events in the Bihar dataset, we employed exponentially weighted moving smoothing to ease the prediction challenge.
Similar to the Milan dataset, we use the initial five days for training (480 samples), the subsequent two days for validation (192 samples), and all remaining data for testing (2829 samples).

\subsection{Baseline Methods}
Multiple state-of-the-art baselines are included in our evaluation to demonstrate the effectiveness of MGSTC both in offline and online forecasting scenarios.

\textbf{Offline Baselines:}
\begin{itemize}
	\item ARIMA \cite{miao2023a}: ARIMA is one of the most classical linear models, which has been widely used for various temporal prediction tasks.
	\item LSTM \cite{ka2023intelligent}: LSTM is a deep learning method based on RNN architectures, which has shown remarkable performance for time series forecasting tasks.
	\item Transformer \cite{vaswani2017attention}: Benefited from the fully self-attention mechanism architecture, Transformer demonstrates exceptional ability in handling long-sequence predictions.
	\item AHSTGNN \cite{wang2023adaptive}: In this method, TCN is employed to extract temporal information, while the GCN and GAT are used to capture static and dynamic spatial correlation respectively, with residual connection applied for final fusion.
	\item MVSTGN \cite{MVSTGN2023Yao}: This model integrates a multi-view attention module to capture global spatial and temporal dependencies, appended by a dense convolution layer to explore local spatial-temporal information.
	\item DCG-MAM \cite{xiao2025cellular}: DCG-MAM serves as a baseline by employing diffusion convolutional GRU and multi-head attention modules to jointly learn spatial, temporal, and periodic dependencies, while also considering external factors in cellular traffic prediction.
\end{itemize}

\textbf{Online Baselines:}
\begin{itemize}
	\item ER \cite{chaudhry2019tiny}: This method stores historical samples in a buffer and periodically interleaves them with new data for model updating.
	\item DER++ \cite{buzzega2020dark}: Originally designed for continual learning, DER++ incorporates a knowledge distillation mechanism on top of the experience replay to address domain shift issues in multi-task learning scenarios.
	\item OneNet \cite{wen2023onenet}: This method consists of two branches: cross-time and cross-variable, with their respective contributions to the prediction task determined by exponential gradient descent and reinforcement learning.
	\item LNTP \cite{zhang2021LNTP}: Wavelet transform is used to decompose the time series data into approximation and detail components, which are then fed into the LSTM model for prediction. A single back-propagation is further performed based on the incoming data for online updating.
	\item DLF \cite{wang2024towards}: DLF provides a baseline approach for dynamic spatial-temporal graph prediction by decoupling time series into seasonal and trend components and using an alternating update strategy to track changes in evolving graphs.
\end{itemize}

\subsection{Experimental Settings}

All experiments are conducted on a desktop with i7-12700K CPU@3.6 GHz processor, 32GB RAM and NVIDIA RTX 4090 GPU. For all datasets, the input historical length is set as $T=128$, and the default prediction horizon is $\tau = 60$. The embedding dimension $D$ is set as 512, and the default chunk and stride lengths are $C=48$, $S=32$. The learning rate is set to be 0.0001 with a batch size of 16 for Milan while 32 for Taiwan and AIIA. The model is trained using the Adam optimizer for 50 epochs, with the early stop if validation set indicator does not drop up to 3 epochs.
For the online learning strategy, the default buffer size $\vert\mathcal{B}\vert$ is 100 samples while the capacity of history repository $\mathcal{H} $ is 256 samples.
For fine-tuning and aggressive update stages, the weight coefficient $\eta^\dagger$ and  $\eta^\ddagger$ are both set as 0.5.
	
Two commonly used metrics are selected for comprehensively evaluation, i.e., Mean Square Error (MSE) and Mean Absolute Error (MAE). MAE is more stable for outliers while MSE is more sensitive to points with larger errors. 
The cumulative average MSEs are also utilized during the test process. The $k$-th cumulative MSE is denoted as $\overline{\textrm{MSE}}_k$, which is defined as the average value of the first $k$ MSEs as
\begin{equation}
	\overline{\textrm{MSE}}_k = \frac{1}{k}\sum^k_{i=1} \textrm{MSE}_i.
\end{equation}
For both metrics, lower values indicate higher prediction accuracy. Additionally, computational complexity, number of parameters and practical runtime are used for efficiency evaluation.

\subsection{Prediction Accuracy}

\begin{table}[t!]
	\caption{MSE and MAE of Various Methods for Cellular Traffic Prediction.}
	\label{tab:acc}
	\begin{threeparttable}
		\resizebox{0.9\columnwidth}{!}{
			\begin{tabular}{c|c|cc|cc|cc|cc}
				\toprule
				& Dataset & \multicolumn{2}{c}{Milan}	& \multicolumn{2}{c}{Taiwan} & \multicolumn{2}{c}{AIIA} & \multicolumn{2}{c}{Bihar}\\
				\midrule
				& Methods & MSE & MAE & MSE & MAE & MSE & MAE & MSE & MAE \\
				\midrule
				\multirow{7}{*}{Offline}& ARIMA & 1.259 & 0.898  & 1.338 & 0.969 & 11.588 & 2.778 & 1.927 & 1.001\\
				& LSTM & 0.836 & 0.659 & 0.479 & 0.531 & 8.859 & 1.975 & 1.217 & 0.834 \\
				& Transformer & 0.761 & 0.592  & 0.564 & 0.528 & 9.454 & 2.019 & 1.283 & 0.843 \\
				& AHSTGNN & OOM & OOM  & 0.659 & 0.601 & 4.766 & 1.899 & 1.101 & 0.811 \\
				& MVSTGN & \underline{0.592} & \underline{0.500} & \underline{0.463} & \underline{0.526} & \underline{0.628} & \underline{0.786} & 1.180 & 0.820 \\
				& DCG-MAM & 0.675 & 0.653 & 0.550 & 0.542 & 1.636 & 1.010 & \underline{1.071} & \underline{0.756}\\
				& \textbf{MGSTC-offline} & \textbf{0.291} & \textbf{0.371} & \textbf{0.359} & \textbf{0.421} & \textbf{0.408} & \textbf{0.408} & \textbf{1.065} & \textbf{0.752} \\
				\cmidrule{2-10}
				& Improvement & 50.8\% & 25.8\% & 22.5\% & 20.0\% & 35.0\% & 48.1\% & 0.5\% & 0.5\% \\
				\midrule
				\midrule
				\multirow{6}{*}{Online}& ER & 0.413 & 0.455 & 0.586 & 0.572 & 0.905 & 0.663 & 1.357 & 0.869 \\
				& DER++ & 0.406 & 0.450 & 0.525 & 0.548 & 0.864 & 0.643 & 1.352 & 0.867 \\
				& OneNet & 0.322 & 0.405 & 0.480 & 0.520 & 0.437 & 0.447 & 1.168 & 0.806 \\
				& LNTP & \underline{0.278} & \underline{0.352} & 0.408 & 0.457 & \underline{0.411} & \underline{0.407} & \underline{1.052} & \underline{0.740} \\
				& DLF & 0.664 & 0.637 & \underline{0.355} & \underline{0.413} & 0.693 & 0.524 & 1.068 & 0.780 \\
				& \textbf{MGSTC} & \textbf{0.266} & \textbf{0.344} & \textbf{0.315} & \textbf{0.403} & \textbf{0.393} & \textbf{0.396} & \textbf{1.049} & \textbf{0.731}\\
				\cmidrule{2-10}
				& Improvement & 4.3\% & 2.3\% & 11.3\% & 2.4\% & 4.4\% & 2.7\% & 0.2\% & 1.2\% \\
				\bottomrule
		\end{tabular}}
		\begin{tablenotes}
			\footnotesize
			\item[1] Optimal values are highlighted in bold, and sub-optimal results are underlined. The ``Improvement'' rows show the performance gain of the optimal method compared to the sub-optimal one.
			\item[2] OOM refers to the out of memory error even if the batch size is set to 1.
		\end{tablenotes}
	\end{threeparttable}
\end{table}

Table~\ref{tab:acc} summarizes the forecasting errors of various methods for both offline and online learning scenarios, where MGSTC-offline refers to the proposed offline network without online learning strategy. The best results are highlighted in bold while sub-optimal results are underlined, and OOM refers to the CUDA out of memory error even if the batch size is set to $1$.
Overall, MGSTC-offline outperforms all offline baseline methods in terms of MSE and MAE. Compared to ARIMA, LSTM and Transformer, MGSTC incorporates the extraction and utilization of spatial correlations, thereby effectively improving prediction accuracy.
While AHSTGNN and MVSTGN integrate spatial information, they focus on short-term historical and prediction horizons and perform exhaustive feature extraction, which hampers their performance on relatively long-term tasks.
DCG-MAM depends on predefined topology input, which introduces subjectivity and lacks adaptability to the four datasets with diverse characteristics.
Furthermore, in the online learning scenario, MGSTC is also superior to other baselines. ER and DER++, as two generic online update strategies, are both rely on experience replay technique without the explicit detection and adaptation for concept drift, thus resulting in large prediction errors.
OneNet tackles the drift issue by ensembling cross-time and cross-variable branches and optimize their weights in real-time. Nevertheless, it also neglects the multi-grained spatial-temporal complementarity of cellular traffic, leading to inadequate performance.
LNTP achieves the sub-optimal accuracy due to its tailored wavelet-LSTM architecture, but fails to induce sufficient parameter changes through single-step update with the latest samples.
DLF, originally designed for transportation traffic flow, is proficient at handling dynamic topological relationships and long-term data evolution but struggles with the fast-varying nature of communication traffic.

\begin{figure}[t]
	\centering
	\subfigure[]{\includegraphics[width=0.462\linewidth, trim=10 8 50 40,clip]{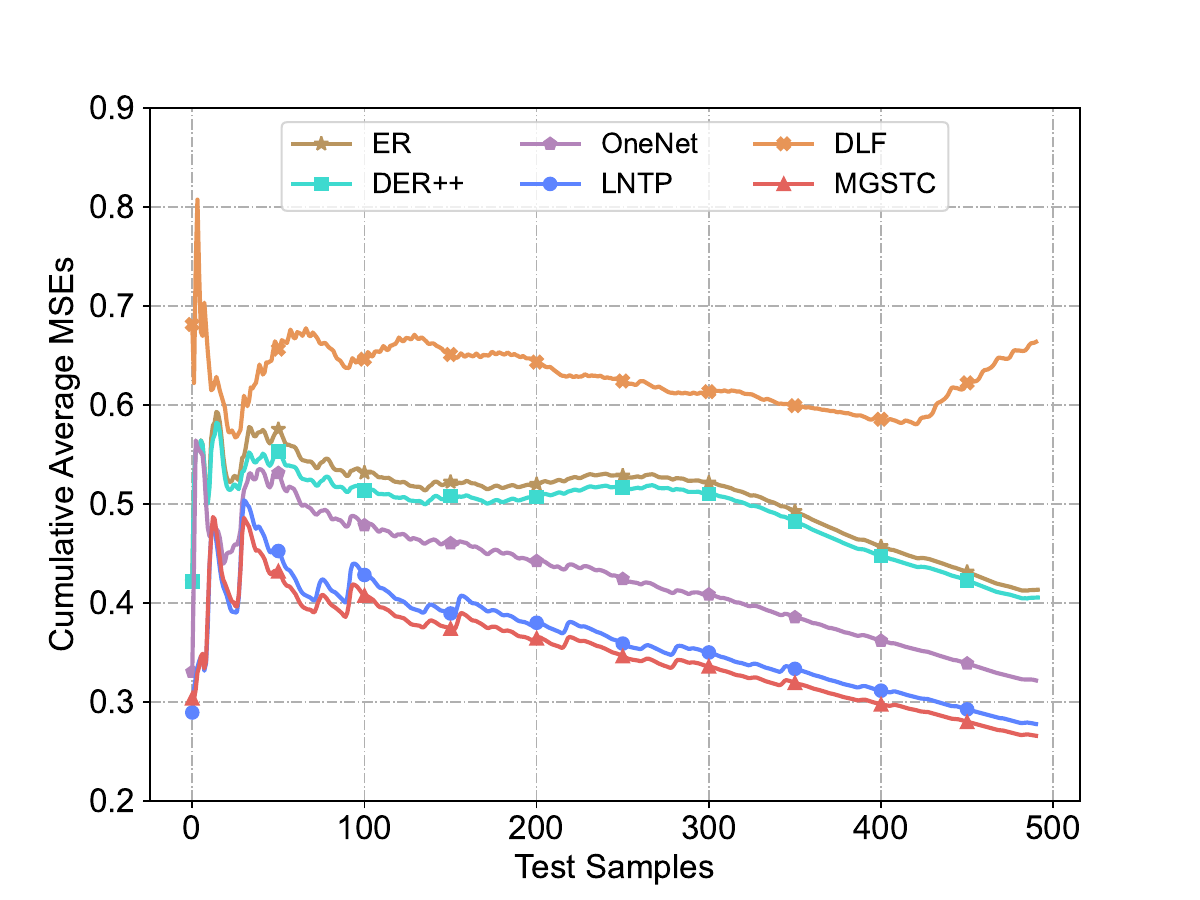} \label{fig:online_baseline_Milan}}
	\vspace{-1em}
	\subfigure[]{\includegraphics[width=0.462\linewidth, trim=10 8 50 40,clip]{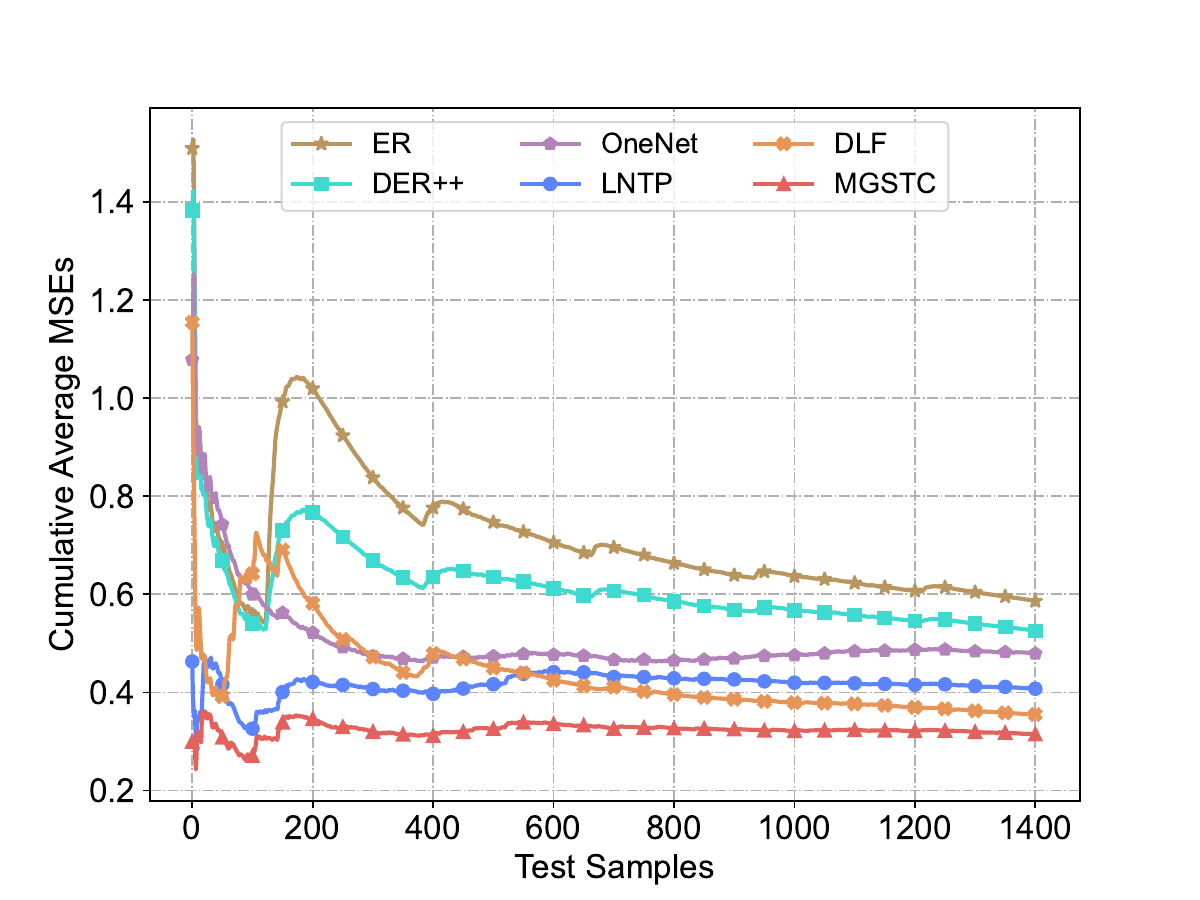} \label{fig:online_baseline_Taiwan}}
	\subfigure[]{\includegraphics[width=0.462\linewidth, trim=10 8 50 40,clip]{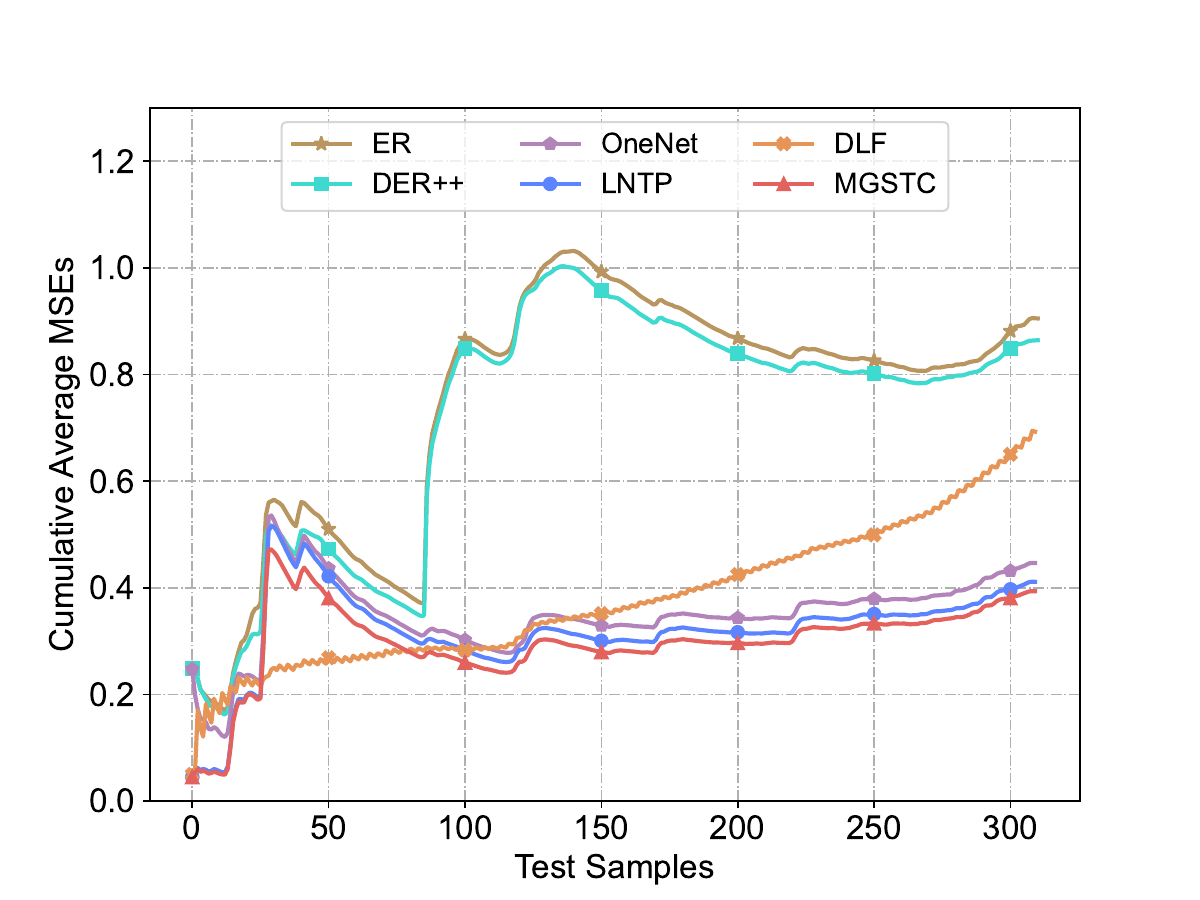} \label{fig:online_baseline_AIIA}}
	\vspace{-1em}
	\subfigure[]{\includegraphics[width=0.462\linewidth, trim=10 8 50 40,clip]{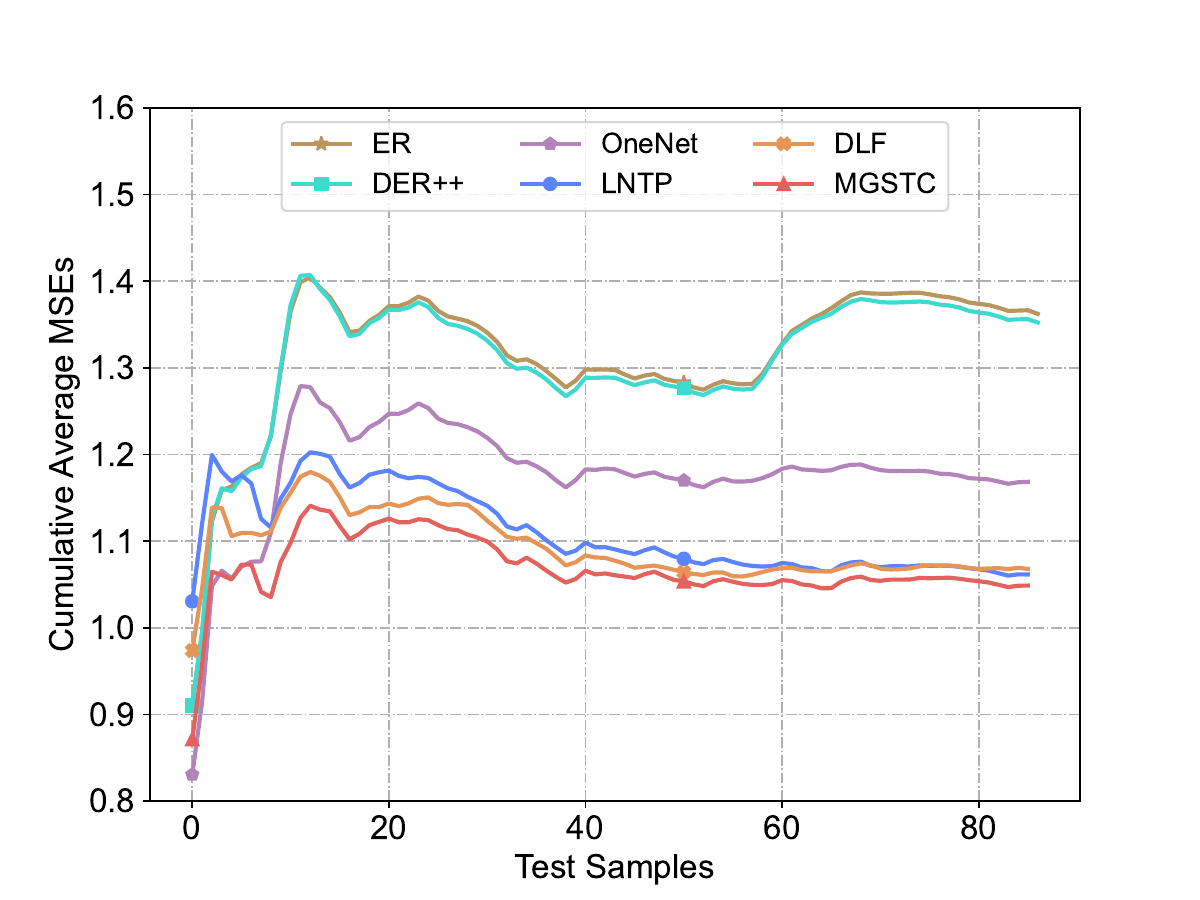} \label{fig:online_baseline_Bihar}}
	\caption{Cumulative average MSEs of various methods on different dataset. (a) Results on Milan dataset. (b) Results on Taiwan dataset. (c) Results on AIIA dataset. (d) Results on Bihar dataset.}
	\label{fig:cumulative_acc}
\end{figure}
We also present the cumulative average MSEs during the test process in Fig. \ref{fig:cumulative_acc}, with markers displayed every 50 points. 
Consistent with the results in Table \ref{tab:acc}, ER and DER++ are susceptible to the drift effects during the updating process, which disrupts the convergence particularly for AIIA. 
We further find the trends of Taiwan and Bihar are stable, while Milan and AIIA show decreasing and increasing trends respectively as the number of test samples grows.
This can be attributed to the unique characteristics of each dataset. In contrast to the Taiwan and Bihar datasets, which shows overall stability, the Milan's traffic volume experiences a decline during the later stages due to holidays, whereas the AIIA shows a significant rise in the later period, leading to the trend differences among four datasets.

\begin{figure}[t]
	\centering
	\subfigure[]{\includegraphics[width=0.462\linewidth, trim=10 8 50 40,clip]{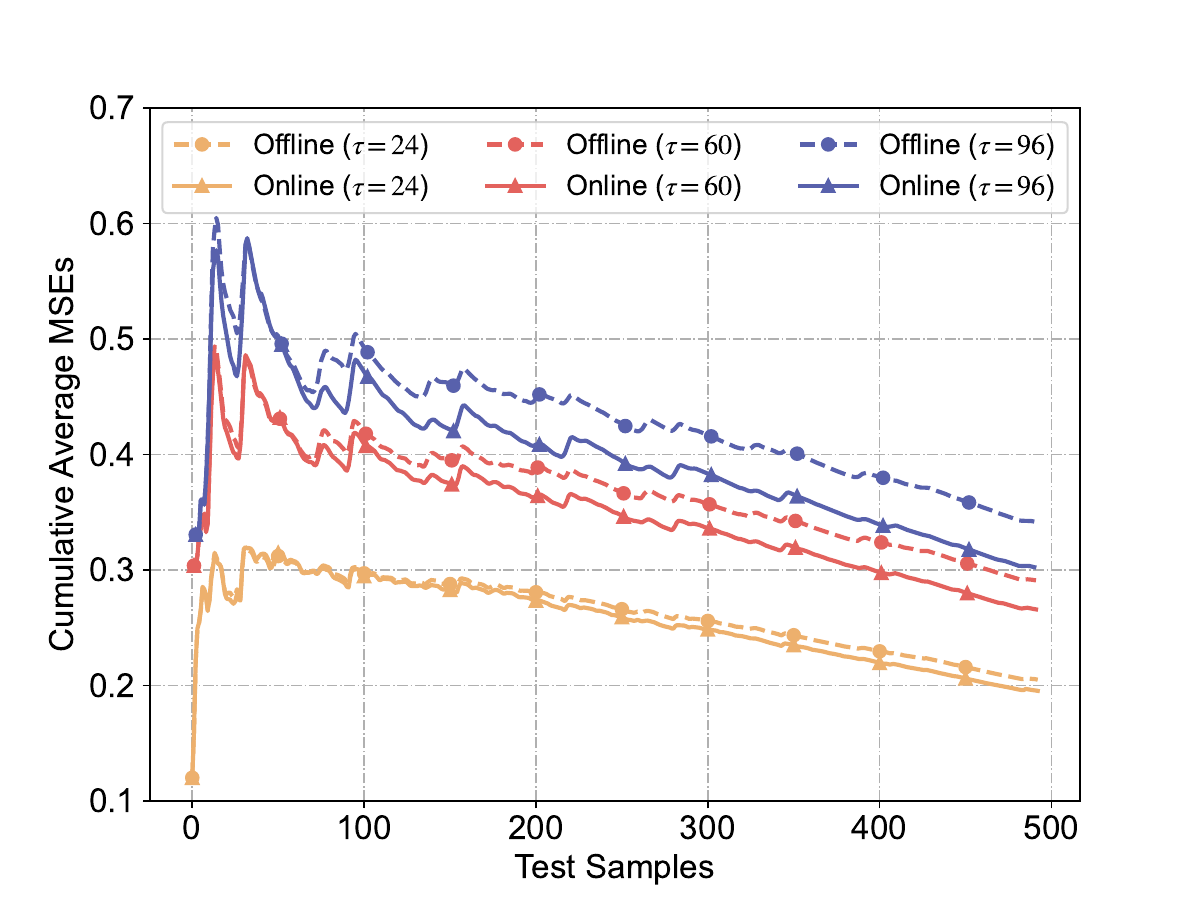} \label{fig:cumulative_offline_online}}
	\subfigure[]{\includegraphics[width=0.462\linewidth, trim=10 8 50 40,clip]{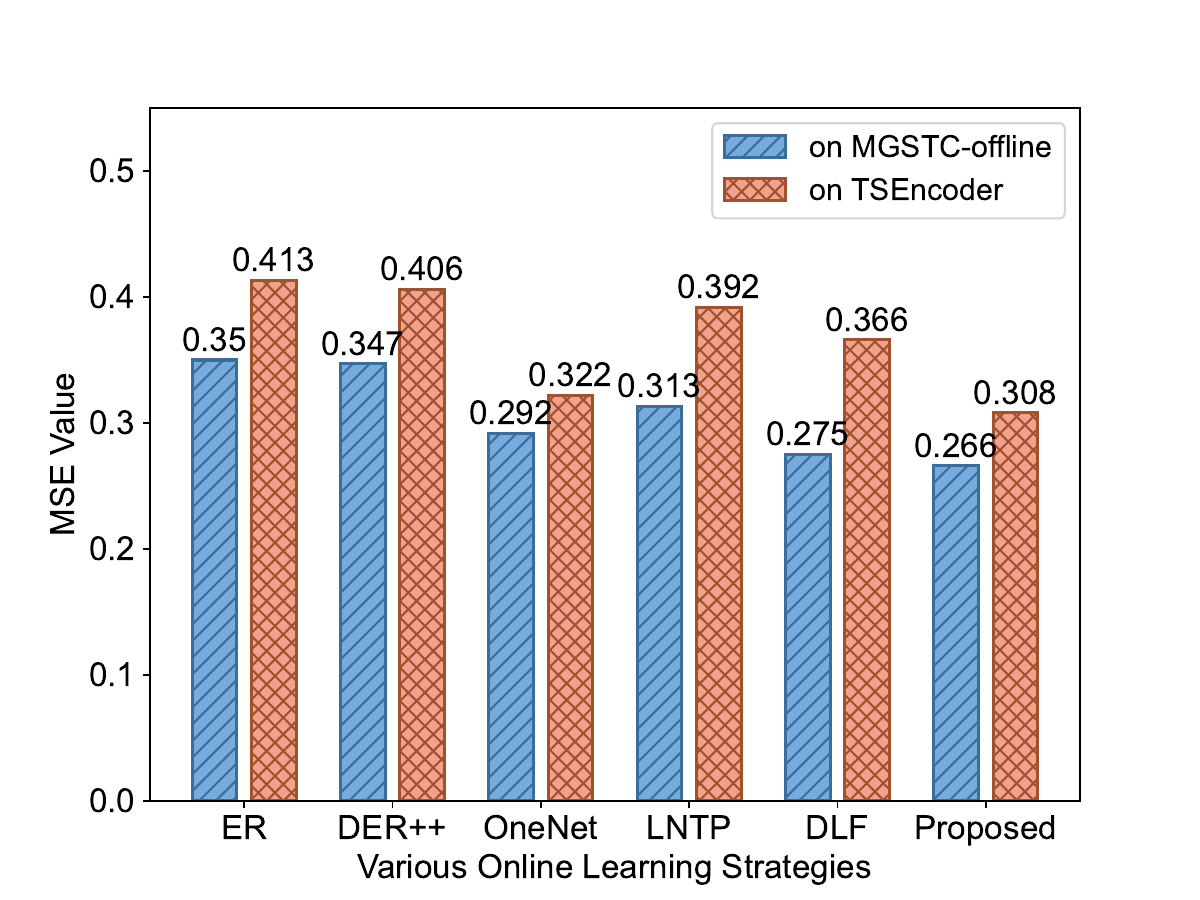} \label{fig:cross_validation_Milan}}
	\vspace{-1em}
	\caption{Effectiveness of online learning strategy. (a) Offline and online cumulative average MSEs of MGSTC with different prediction lengths. (b) MSE performance of various online learning strategies on the same offline models.}
	\label{fig:online}
\end{figure}
Furthermore, we demonstrate the effectiveness of the proposed online learning strategy through Fig.~\ref{fig:online}.
The comparison of $\overline{\textrm{MSE}}_k$ in both offline and online scenarios is depicted in Fig.~\ref{fig:cumulative_offline_online}. We can clearly find that online approach outperforms the offline approach for all prediction lengths $\tau$. More specifically, slight fluctuations can be observed during the testing process, suggesting the occurrence of concept drift. In this context, MGSTC exhibits faster adaptation to the new characteristics than MGSTC-offline, gradually accumulating an advantage. 
Moreover, we also test the performance of different online methods on two unified offline models, namely MGSTC-offline and TSEncoder. TSEncoder is the offline model of ER, DER++ and OneNet, which is built with dilated convolutions. From Fig.~\ref{fig:cross_validation_Milan}, we can observe that the proposed online learning strategy attains optimal results and MGSTC-offline exhibits superior fitting ability compared to TSEncoder.

\subsection{Computation Efficiency}

Oriented towards the requirements of extremely low latency in future mobile networks \cite{liu2023machine}, operational efficiency is a critical consideration when deploying algorithms in the practical communication infrastructures.
Hence, computation efficiency is also analyzed at both theoretical and practical level. Table \ref{tab:complexity} summarizes the computational complexity and number of parameters of various methods, where $p$ and $q$ indicate the order of auto-regressive and moving average of ARIMA, and $E$ denotes the number of edges in a graph. $P$ denotes the number of diffusion steps of DCG-MAM and $n$ is the number of sampled points during the online process of DLF. We can find that MGSTC's complexity and parameter count are relatively small. This is attributed to MGSTC's use of chunking for long sequences, enabling a considerable reduction in time complexity by a factor of $1/C^2$. Moreover, the complexity of the FGSA is only linearly related to $T$ and $D$, which is significantly lower than the GNNs' complexity quadratic with $D$.
\begin{table}[h]
	\caption{Computational complexity and Number of Parameters of Various Methods.}
	\label{tab:complexity}
	\centering
	\resizebox{1.0\columnwidth}{!}{
		\begin{threeparttable}
			\begin{tabular}{ccc|ccc}
				\toprule
				Offline Methods & Complexity & Parameters & Online Methods & Complexity & Parameters \\
				\midrule
				ARIMA & $\mathcal{O}(T(p+q))$ & 0.01M & ER & $\mathcal{O}(T^2D+T)$ & 18.03M \\
				LSTM & $\mathcal{O}(TD^2)$ & 0.02M & DER++ & $\mathcal{O}(T^2D+T)$ & 18.03M \\
				Transformer & $\mathcal{O}(T^2D)$ & 13.74M & OneNet & $\mathcal{O}(T^2D+ND^2)$ & 19.45M \\
				AHSTGNN & $\mathcal{O}((T+N)D^2+ED)$ & 0.87M & LNTP & $\mathcal{O}(T^2 + TD^2)$ & 0.30M \\
				MVSTGN & $\mathcal{O}((T+N)D^2+ED)$ & 0.54M & DLF & $\mathcal{O}(TN^2D+Tn^2D)$ & 0.07M \\
				DCG-MAM & $\mathcal{O}(TP(N+D)+TD^2)$ & 0.02M & \textbf{MGSTC} & $\mathcal{O}(T^2D/C^2+ND)$ & 0.50M \\
				\bottomrule
			\end{tabular}
			\begin{tablenotes}
				\footnotesize
				\item[1] $p$ and $q$ denote the order of auto-regressive and moving average, respectively. $E$ is the number of edges for a graph. $P$ denotes the number of diffusion steps of DCG-MAM. $n$ is the number of sampled points during the online process of DLF.
			\end{tablenotes}
	\end{threeparttable}}
\end{table}
\begin{figure}[h]
	\centering
	\subfigure[]{\includegraphics[width=0.485\linewidth, trim=5 8 5 0,clip]{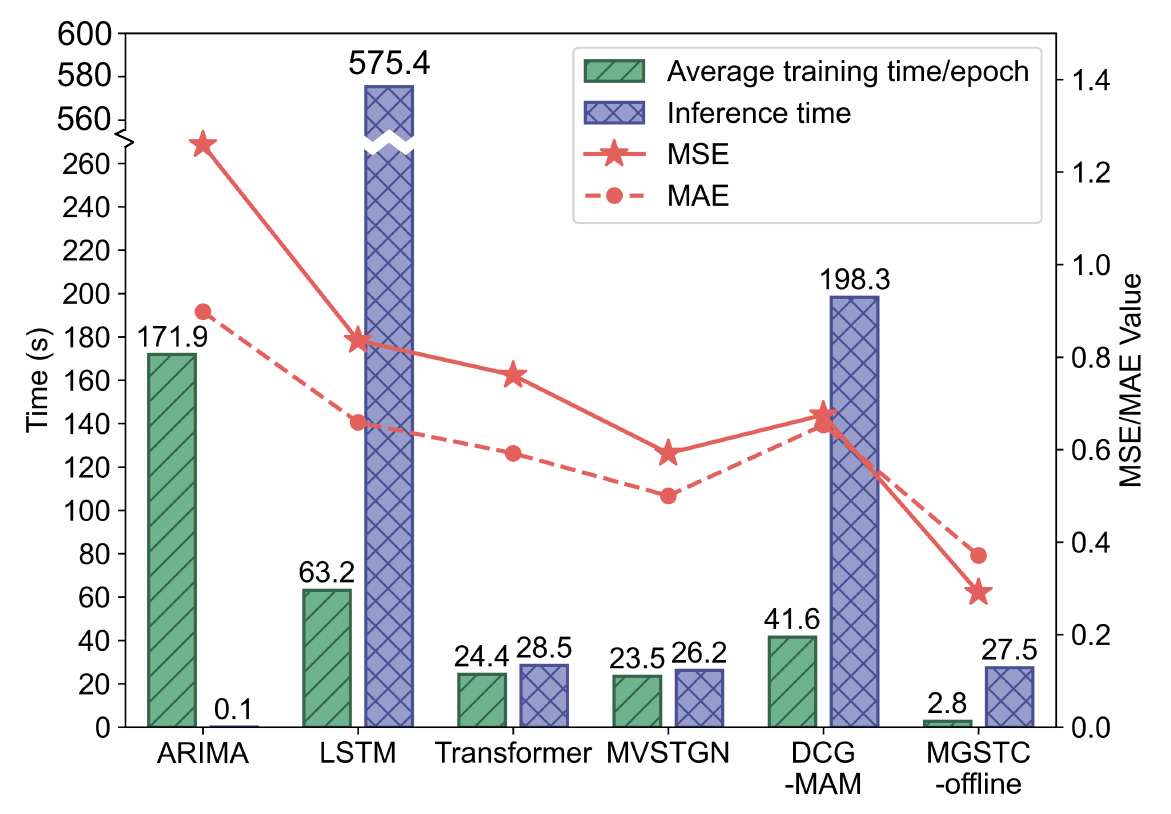} \label{fig:offline_efficiency}}
	\subfigure[]{\includegraphics[width=0.49\linewidth, trim=20 20 10 10,clip]{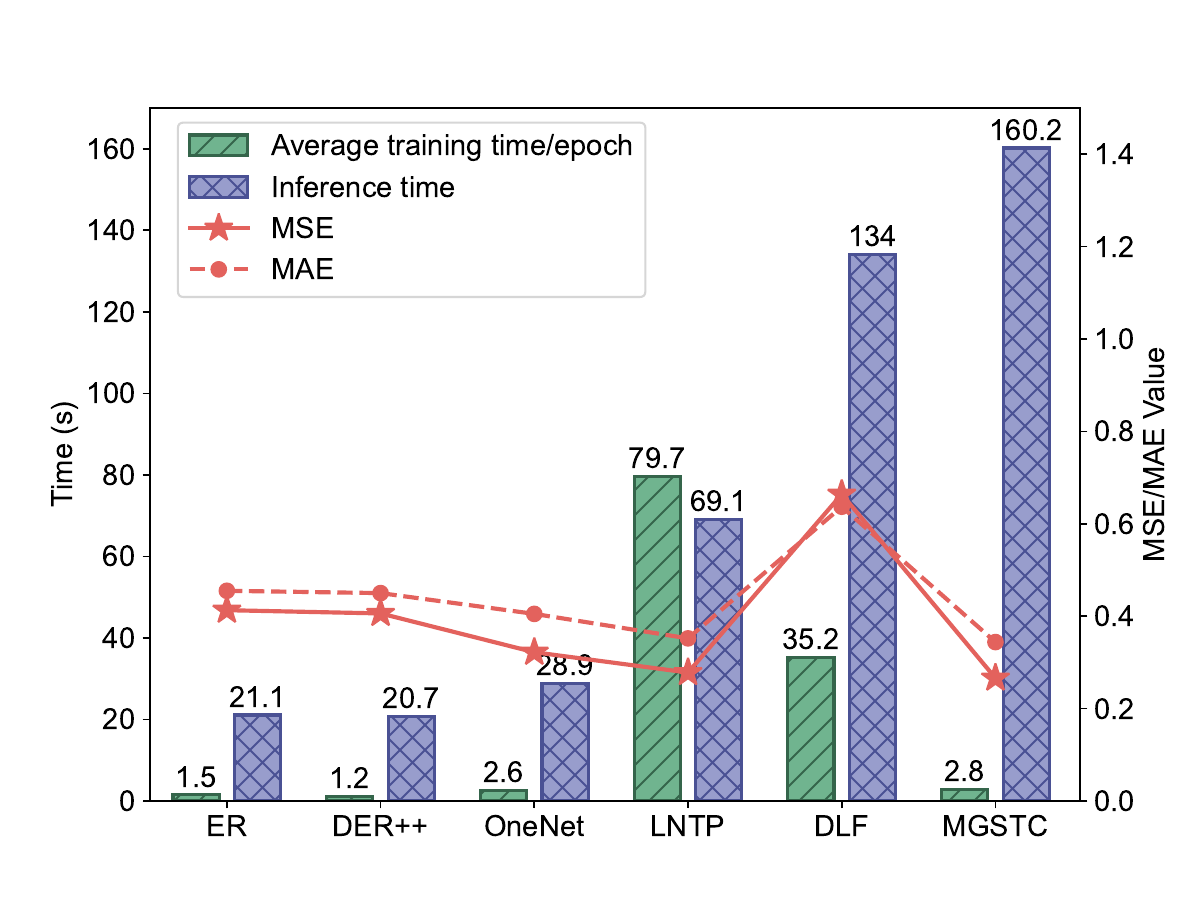} \label{fig:online_efficiency}}
	\caption{Average training time/epoch and inference time of various methods on Milan dataset. (a) Offline runtime comparison. (b) Online runtime comparison.}
	\label{fig:efficiency}
\end{figure}

Additionally, practical runtime including the average training time for each epoch and the inference time of various methods on Milan dataset are illustrated in Fig.~\ref{fig:efficiency}, except for the AHSTGNN which incurs OOM error. The MSE and MAE metrics for each method are also presented using line charts. Fig.~\ref{fig:offline_efficiency} shows that despite the low complexity of ARIMA and LSTM, their training process is time-consuming due to their inability for parallel training. In contrast, self-attention-based methods significantly reduce training time through parallelization, while MGSTC further accelerates training by chunks. For online learning scenario, Fig.~\ref{fig:online_efficiency} shows that MGSTC takes the longest inference time due to its exclusive aggressive update module, which allows MGSTC to adapt to the cellular concept drift in real time, thereby boosting prediction accuracy.
Given the 10-minute sampling interval of the Milan dataset, MGSTC is able to forecast traffic for the upcoming 10 hours in around 160 seconds. Despite the moderate efficiency, the time consumption is reasonable and completely acceptable.

\subsection{Impact of Model Parameters}

\begin{figure}[h]
	\centering
	\subfigure[]{\includegraphics[width=0.28\linewidth]{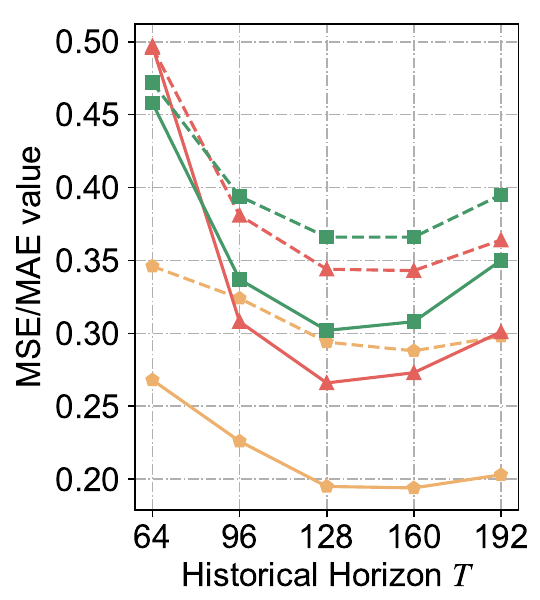} \label{impact_seqlen}}
	\subfigure[]{\includegraphics[width=0.28\linewidth]{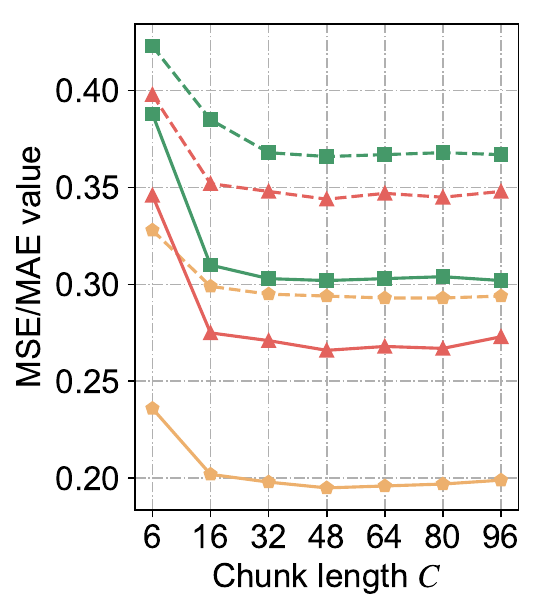} \label{impact_chunklen}}
	\subfigure[]{\includegraphics[width=0.28\linewidth]{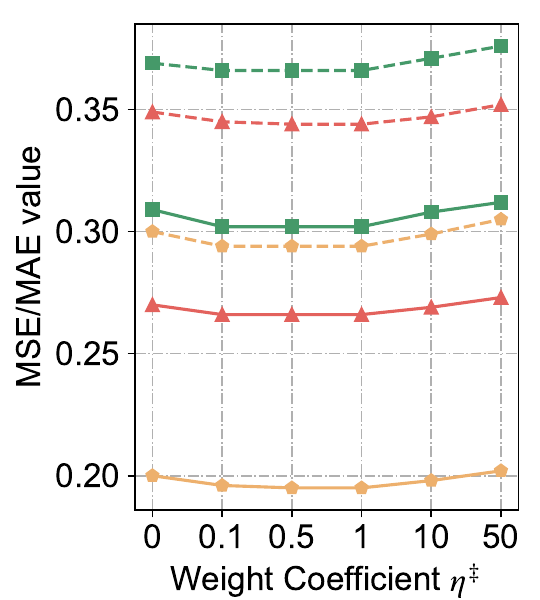} \label{impact_eta}}
	\vspace{-1em}
	\subfigure[]{\includegraphics[width=0.278\linewidth, trim=-15 0 0 0, clip]{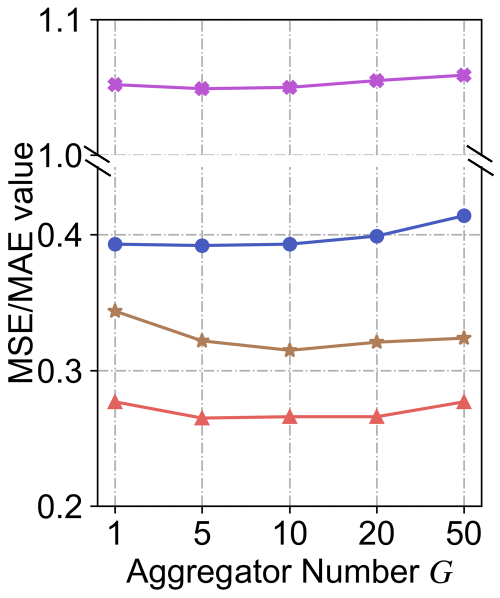} \label{impact_router}}
	\subfigure[]{\includegraphics[width=0.289\linewidth, trim=-15 0 0 0, clip]{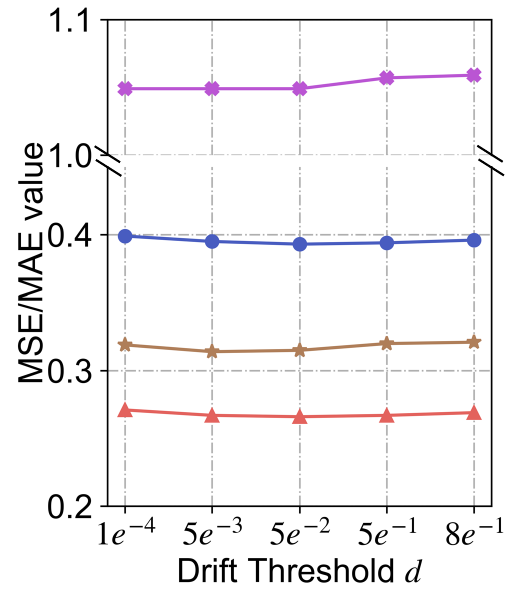} \label{impact_d}}
	\subfigure[]{\includegraphics[width=0.275\linewidth, trim=-15 0 0 0, clip]{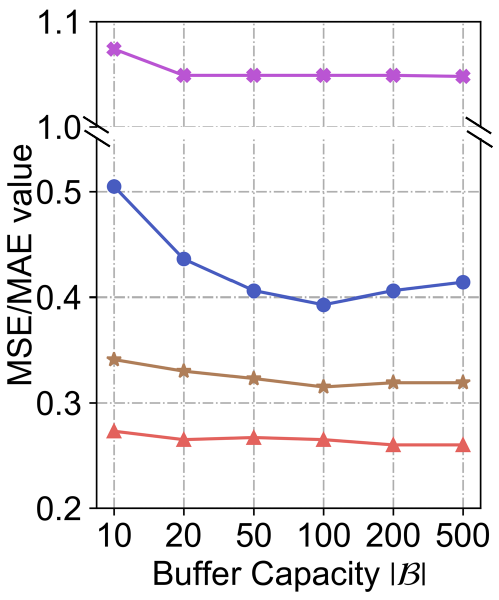} \label{impact_buffer}}
	\subfigure{\includegraphics[width=1.0\linewidth, trim=0 0 0 0, clip]{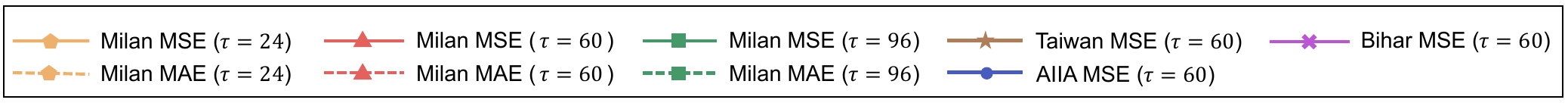} \label{impact_legend}}
	\caption{Impact of model parameters. (a) Impact of historical horizon $T$. (b) Impact of chunk length $C$. (c) Impact of aggressive update coefficient $\eta^\ddagger$. (d) Impact of aggregator number $G$. (e) Impact of drift threshold $d$. (f) Impact of buffer capacity $\vert\mathcal{B}\vert$}.
	\label{fig:impact}
\end{figure}

In this subsection, we adjust several key model parameters to reveal their impact on prediction performance as shown in Fig. \ref{fig:impact}. 

We can observe from Fig.~\ref{impact_seqlen} that longer input historical sequence is not always beneficial, as data from distant past tends to have weaker correlations and distracts the model's attention, causing a decline in performance. Figure \ref{impact_chunklen} illustrates that overly small chunk lengths degrade model accuracy due to the propagation of occasional noise caused by excessively fine-grained temporal exploration. Conversely, longer chunks result in only trivial increase in MSE and MAE, since the complementary nature of spatial-temporal information ensures that coarse-grained temporal features can be compensated by refined spatial granularity. Fig.~\ref{impact_eta} demonstrates that the prediction accuracy is robust to aggressive update coefficient $\eta^\ddagger$ ranging from $0$ to $50$. The minimum error is observed around 0.5, implying the most suitable proportion of historical data's participation.

As shown in Fig.~\ref{impact_router}, the impact of number of aggregator $G$ is investigated on four datasets. This parameter depends on the number of series in the dataset, i.e., fewer series require fewer aggregators. Specifically, MGSTC performs best on Milan dataset with $5$ to $20$ aggregators, Taiwan and Bihar prefer $G = [5, 10]$, and AIIA favors even fewer aggregators ($G<10$).
Regarding the concept drift threshold $d$, smaller threshold means the stricter detection standard and the fewer aggressive update process. Fig.~\ref{impact_d} reveals that excessively frequent aggressive updates fail to enhance prediction accuracy. A moderate threshold $d$ around $5e^{-2}$ proves beneficial for Milan, AIIA and Bihar whereas a threshold around $5e^{-3}$ works better for Taiwan. The impact of buffer capacity $\vert \mathcal{B} \vert$ is also illustrated in Fig.~\ref{impact_buffer}. Our results show that an insufficient buffer capacity causes varying levels of MSE degradation on all four datasets. This occurs because a small buffer is more likely to be influenced by short-term noise, making it unreliable in representing the overall traffic trend, which in turn can result in incorrect concept drift detection. Moreover, while prediction errors remain relatively stable with larger $\vert \mathcal{B} \vert$, it comes at the cost of increased storage consumption.

\subsection{Ablation Experiment}

The investigation into the necessity of each component of MGSTC involves a systematic exclusion of components to observe the corresponding reduction in model performance. The experiments are conducted on all datasets, and the ablation experiment results are shown in Table~\ref{tab:ablation}. 
\begin{table}[h]
	\caption{Ablation Results on Four Datasets.}
	\label{tab:ablation}
	\resizebox{1.0\columnwidth}{!}{
		\begin{tabular}{p{6cm}|cc|cc|cc|cc}
			\toprule
			\multicolumn{1}{c}{Datasets} & \multicolumn{2}{c}{Milan} & \multicolumn{2}{c}{Taiwan} & \multicolumn{2}{c}{AIIA} & \multicolumn{2}{c}{Bihar}\\
			\midrule
			Methods & MSE & MAE & MSE & MAE & MSE & MAE & MSE & MAE \\
			\midrule
			MGSTC w/o Chunk &OOM &OOM & 0.459& 0.493&0.681 &0.588 & 1.180 & 0.783\\
			MGSTC w/o FGSA &0.277 &0.350 & 0.408 & 0.457 & 0.425& 0.416 & 1.062 & 0.738 \\
			MGSTC w/o Short-term memory ($\eta^\dagger=0$) & 0.275& 0.357& 0.332& 0.415& 0.398& 0.399 & 1.061 & 0.734 \\
			MGSTC w/o Aggressive Update &0.272 & 0.352 & 0.325& 0.411 & 0.401 & 0.403 & 1.064 & 0.735\\
			MGSTC w/o Long-term memory ($\eta^\ddagger=0$) & 0.270 & 0.349 & 0.323&0.410 & 0.396 & 0.397 & 1.059 & 0.733 \\
			MGSTC w/o perturbation ($p=0$) & 0.268& 0.347&0.319 &0.405 & 0.395 & 0.397 & 1.052 & 0.732 \\
			MGSTC & 0.266 & 0.344 & 0.315 & 0.403 & 0.393 & 0.396 & 1.049 & 0.731\\
			\bottomrule
	\end{tabular}}
\end{table}

Compared with the original MGSTC, removing chunks results in the most severe performance degradation and run out of memory on Milan dataset, showing that overly precise treatment of temporal dependencies can have adverse effects. The considerable gap between MGSTC and MGSTC w/o FGSA shows that spatial correlations are also an indispensable source of information for improving prediction accuracy. Moreover, MGSTC w/o short-term memory means that sample $(\bar{\mathbf{X}}^{(t-T+1:t)}, \bar{\mathbf{Y}}^{(t+1:t+\tau)})$ is excluded during the fine-tuning process, i.e., $\eta^\dagger=0$, and the results highlight the necessity of buffer $\mathcal{B}$ during fine-tuning. The entire removal of the aggressive update implies the inability to detect and adapt to concept drift, which in turn causes a decrease in accuracy. Moreover, the absence of historical information from $\mathcal{H}$ in the aggressive update, denoted by MGSTC w/o long-term memory,  weakens the adaptation to concept drift. While the removal of perturbations $p$ is marginally better than MGSTC w/o long-term memory, it still shows a slight drop, confirming the positive impact of augmentation on concept drift adaptation. 

\subsection{Case Study}

\begin{figure}[h]
	\centering
	\subfigure[]{\includegraphics[width=0.49\linewidth]{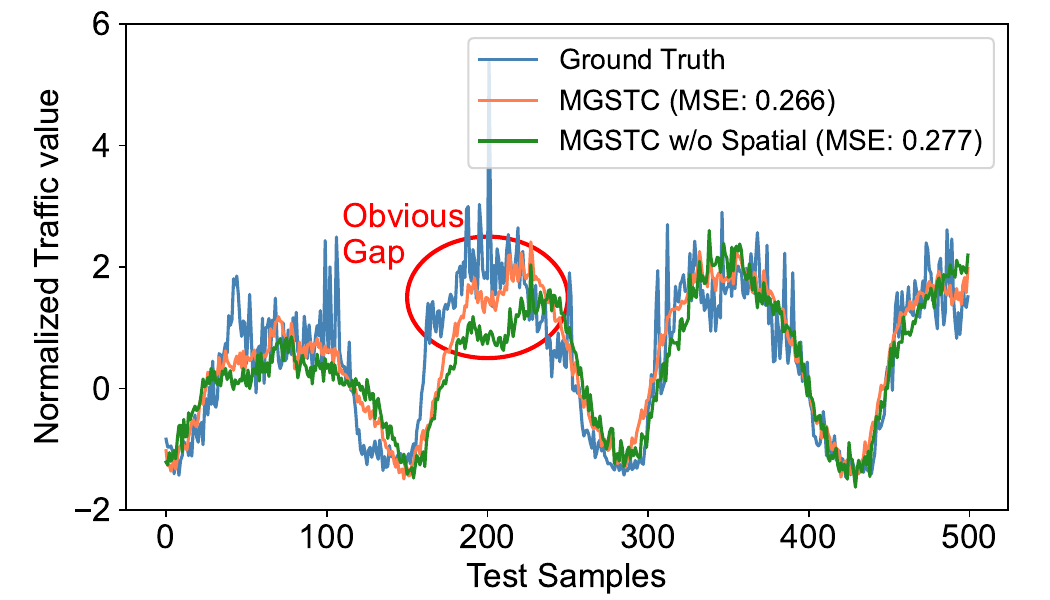} \label{case_Milan_spatial}}
	\subfigure[]{\includegraphics[width=0.48\linewidth]{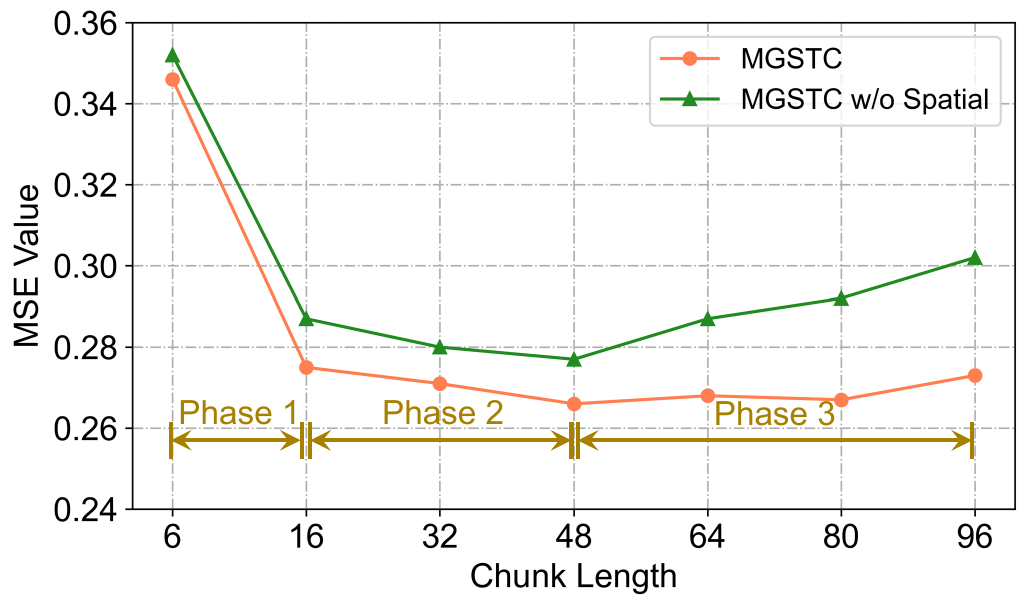} \label{case_Milan_spatial_MSE}}
	\caption{Visualization of multi-grained spatial-temporal feature complementarity on Milan dataset. (a) Comparison of predicted value and ground truth for both MGSTC and MGSTC w/o spatial information. (b) MSE performance of MGSTC and MGSTC w/o spatial information versus chunk length.}
	\label{fig:case_spatial}
\end{figure}

This subsection presents a case study based on Milan dataset to illustrate the effects of multi-grained spatial-temporal feature complementarity and the proposed online learning strategy. Fig. \ref{fig:case_spatial}\subref{case_Milan_spatial} visualizes a 25-step-ahead prediction segment extracted from the third node of the Milan dataset. We can observe that MGSTC achieves better alignment with the ground truth compared to MGSTC w/o spatial information, especially around 200 timestamps. Furthermore, we depict the MSE performance of MGSTC and MGSTC w/o spatial information under varying chunk lengths in Fig. \ref{fig:case_spatial}\subref{case_Milan_spatial_MSE}, which can be divided into three phases. In the first phase, the MSE drops sharply as the chunk length increases, as moderately larger chunk lengths help suppress the propagation of incidental noise or fluctuations. During the second phase, both approaches are able to consistently sustain low prediction errors, indicating their robustness across a range of chunk lengths. In the third phase, the error of MGSTC starts to rise gradually, while the MSE of MGSTC w/o spatial information increases more rapidly.
This implies that the deficiencies of temporal feature extraction can be partially compensated by spatial features.
Benefiting from the complementarity of spatial-temporal features, MGSTC not only achieves superior performance than MGSTC w/o spatial information, but also maintains robustness across a wider range of chunk sizes.

\begin{figure}[h]
	\centering
	\subfigure[]{\includegraphics[width=0.49\linewidth]{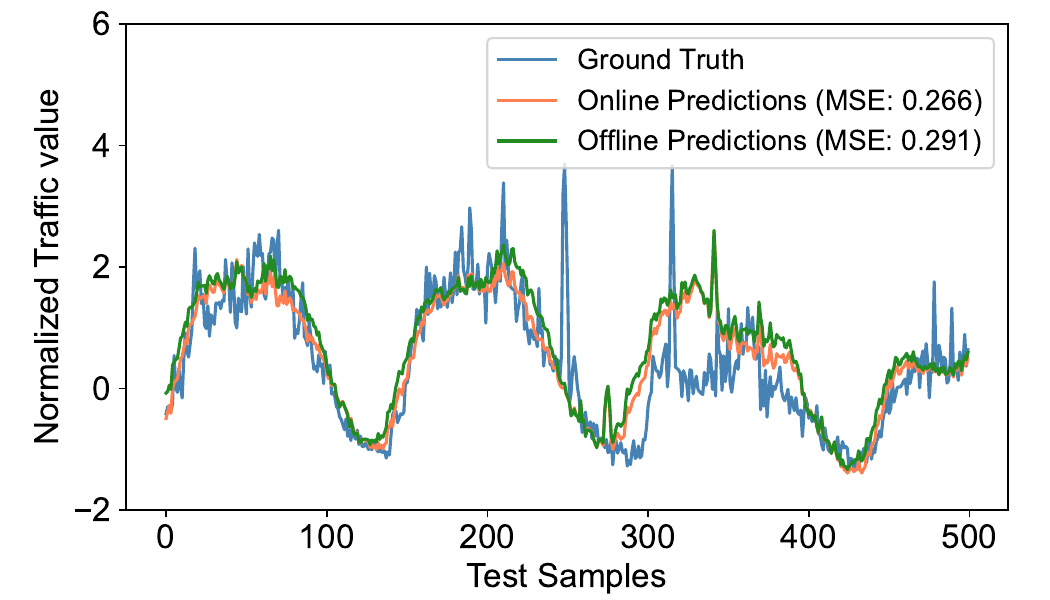} \label{case_Milan_online}}
	\subfigure[]{\includegraphics[width=0.49\linewidth]{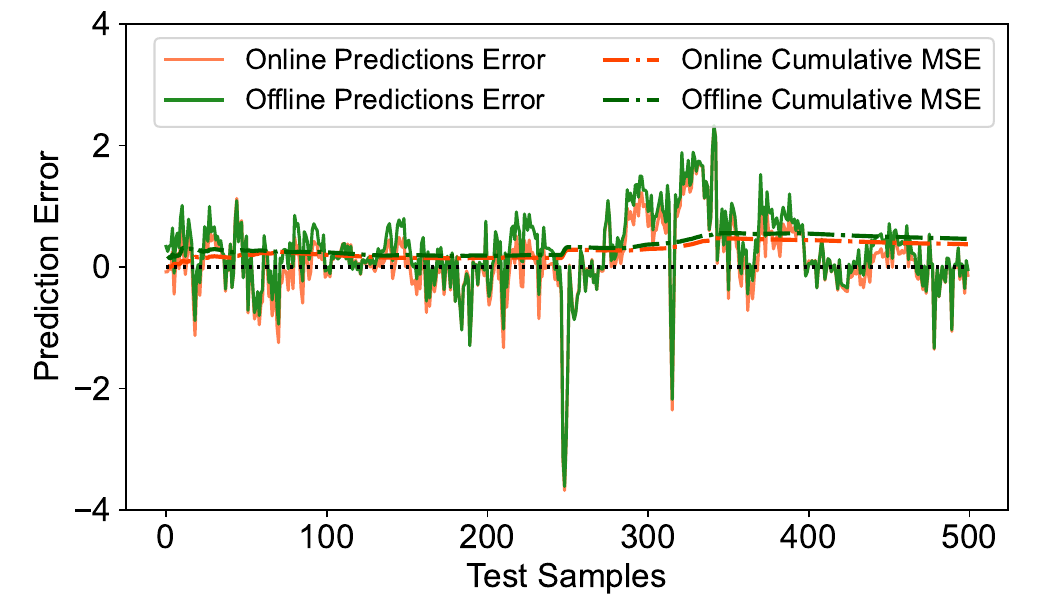} \label{case_Milan_online_error}}
	\caption{Visualization of offline and online predictive methods on Milan dataset. (a) Comparison of predicted value and ground truth for both MGSTC and MGSTC-offline. (b) Step-wise error and cumulative MSE for both MGSTC and MGSTC-offline.}
	\label{fig:case_online}
\end{figure}

Additionally, the effectiveness of the online learning strategy is illustrated in Fig. \ref{fig:case_online}. The predictive results of MGSTC and MGSTC-offline are visualized in Fig.~\ref{fig:case_online}\subref{case_Milan_online} while the step-wise error and cumulative MSE are presented in \ref{fig:case_online}\subref{case_Milan_online_error}. 
We can find that MGSTC can better adapt to new data patterns through fine-tuning and aggressive update, thereby achieving lower cumulative MSE.
This case study highlights the remarkable adaptability of MGSTC to dynamic and evolving data patterns.

\section{Conclusion}

In this paper, we investigate the online cellular traffic prediction problem. Based on the in-depth analysis of cellular traffic characteristics, we develop a novel prediction method MGSTC. Specifically, MGSTC segments the input long sequences into chunks, and employs the CGTA to capture the coarse-grained temporal dependencies, which provide the trend reference for prediction window. Then the constructed trend is refined through the detailed spatial correlations extracted by FGSA. The CGTA and FGSA work in a complementary manner to disseminate valuable information effectively.
Furthermore, to accommodate the continuous forecasting requirement, we enhance the predictive network with an online learning strategy, which enables MGSTC to monitor and adapt to the cellular concept drift in real-time.
In comparison to baseline methods, MGSTC demonstrates superior performance in both offline and online scenarios. The experimental results also provide valuable insights into the impact of key model parameters, while the ablation study validates the effectiveness of each component. Given that this paper primarily focuses on accuracy improvement, its computational efficiency remains to be optimized. In future work, we aim to develop lightweight forecasting models to enhance the feasibility of real-world deployment.

\section*{Acknowledgments}

This work was supported in part by the National Science and Technology Major Project under Grant No.~2024ZD1300200 and the Fundamental Research Funds for the Central Universities under Grant No.~2242023R40005. We also wish to thank Dr. Huazhou Hou at PML for helpful discussions.

\section*{Appendix: Proof of the data augmentation's effectiveness}

The total streaming cellular traffic data can be classified into two categories: historical data $\mathcal{D}_A$ and newly arrived data $\mathcal{D}_B$. The complete data distribution $\mathcal{D}$ can be articulated as a linear combination of $\mathcal{D}_A$ and $\mathcal{D}_B$: $\mathcal{D} = (1-\gamma)\mathcal{D}_A + \gamma \mathcal{D}_B$, where $\gamma \in (0, 1)$ is a fixed weight coefficient.
Simplifying the input historical traffic as $\mathbf{x}$ and the ground truth results as $\mathbf{y}$, we reformulate the cellular forecasting task as an optimization problem:
\begin{equation}
	\min_{\mathbf{w}\in\mathbb{R}^d} \mathcal{L}(\mathbf{w}) = \mathbb{E}_{\mathbf{x}\sim \mathcal{D}}\left[\vert \mathbf{y} - \langle \mathbf{x}, \mathbf{w} \rangle \vert ^2\right],
\end{equation}
where $\mathbf{w}$ represents the learnable parameters of MGSTC, and the optimal parameters $\mathbf{w}^\ast$ can be derived by
\begin{equation}
	\begin{aligned}
	\mathbf{w}^\ast &= \mathbf{R}^{-1}\mathbb{E}_{\mathbf{x}\sim \mathcal{D}}[\mathbf{y}\mathbf{x}] \\
		& = \mathbf{R}^{-1}\left(\mathbb{E}_{\mathbf{x}\sim \mathcal{D}_A}[\mathbf{y}\mathbf{x}] + \mathbb{E}_{\mathbf{x}\sim \mathcal{D}_B}[\mathbf{y}\mathbf{x}] \right),
	\end{aligned}
\end{equation}
where $\mathbf{R}$ is the auto-correlation matrix of $\mathbf{x}$, which is defined as $\mathbf{R} = \mathbb{E}_{\mathbf{x}\sim \mathcal{D}}[\mathbf{x}\mathbf{x}^\mathsf{T}]$.
However, in typical prediction tasks, we only have access to the historical training data $\mathcal{D}_A$, and can achieve the sub-optimal prediction models composed of parameters $\mathbf{w}_A$,
\begin{equation}
	\mathbf{w}_A = \mathbf{R}_A^{-1} \mathbb{E}_{\mathbf{x}\sim \mathcal{D}_A}[\mathbf{y}\mathbf{x}],
\end{equation}
where $\mathbf{R}_A = \mathbb{E}_{\mathbf{x}\sim \mathcal{D}_A}[\mathbf{x}\mathbf{x}^\mathsf{T}]$.
Next, we will examine the gap between $\mathbf{w}^\ast$ and $\mathbf{w}_A$. Since the amount of historical training data $\mathcal{D}_A$ is much larger than test data $\mathcal{D}_B$, $\mathbb{E}_{\mathbf{x}\sim \mathcal{D}_A}[\mathbf{y}\mathbf{x}] + \mathbb{E}_{\mathbf{x}\sim \mathcal{D}_B}[\mathbf{y}\mathbf{x}]$ can be approximated by $\mathbb{E}_{\mathbf{x}\sim \mathcal{D}_A}[\mathbf{y}\mathbf{x}]$. Hence, the primary discrepancy between $\mathbf{w}^\ast$ and $\mathbf{w}_A$ is determined by the gap between $\mathbf{R}$ and $\mathbf{R}_A$.
Furthermore, considering the large volume of $\mathcal{D}_A$, we suppose that $K$ groups of features are strongly correlated in $\mathcal{D}_A$, where $K<T$. Therefore, $\mathbf{R}_A$ can be approximated as
\begin{equation}
	\mathbf{R}_A = \alpha\mathbf{I} + \mathbf{U} \mathrm{diag}(\boldsymbol{\nu})\mathbf{U}^\mathsf{T},
	\label{eq:Z_A}
\end{equation}
where $\alpha >0$, $\mathbf{I}$ is a identity matrix, $\mathbf{U} \in \mathbb{R}^{T\times K}$ is an orthonormal matrix and $\boldsymbol{\nu} \in \mathbb{R}^K_+$. Additionally, it can be assumed that most features in $\mathcal{D}_B$ are uncorrelated, and thus $\mathbf{R}_B$ can be approximated by
\begin{equation}
	\mathbf{R}_B = \beta \mathbf{I},
	\label{eq:Z_B}
\end{equation}
where $\beta >0$.

To quantitatively measure the difference, a gap function $G(\mathbf{R}_A)$ is defined as
\begin{equation}
	\begin{aligned}
		G(\mathbf{R}_A) & = \mathbf{R}^{-1/2}(\mathbf{R}-\mathbf{R}_A)\mathbf{R}^{-1/2} \\
		& = \gamma \mathbf{R}^{-1/2}(\mathbf{R}_B-\mathbf{R}_A)\mathbf{R}^{-1/2}.
	\end{aligned}
\end{equation}
By substituting the expressions for $\mathbf{R}_A$ in \eqref{eq:Z_A} and $\mathbf{R}_B$ in \eqref{eq:Z_B}, $G(\mathbf{R}_A)$ can be reformulated as
\begin{equation}
	\begin{aligned}
		G(\mathbf{R}_A) &= \gamma (\lambda\mathbf{I} +  \gamma\mathbf{U}\mathrm{diag}(\boldsymbol{\nu})\mathbf{U}^\mathsf{T})^{-1/2} ((\beta-\alpha)\mathbf{I} - \mathbf{U}\mathrm{diag}(\boldsymbol{\nu})\mathbf{U}^\mathsf{T})(\lambda\mathbf{I} +  \gamma\mathbf{U}\mathrm{diag}(\boldsymbol{\nu})\mathbf{U}^\mathsf{T})^{-1/2} \\
		&= \gamma(\beta-\alpha+\frac{\lambda}{\gamma}) (\lambda\mathbf{I} + \gamma  \mathbf{U}\mathrm{diag}(\boldsymbol{\nu})\mathbf{U}^\mathsf{T})^{-1} -\mathbf{I},
	\end{aligned}
\end{equation}
where $\lambda = (1-\gamma)\alpha + \gamma \beta$. Given that $\beta >\alpha$, the L2 norm of $G(\mathbf{R}_A)$ can be obtained as
\begin{equation}
	\vert G(\mathbf{R}_A)\vert_2 = \max\left(\frac{\gamma(\beta-\alpha)}{\lambda}, 1 - \frac{\gamma(\beta-\alpha)+\lambda}{\lambda + \gamma\vert \boldsymbol{\nu}\vert_\infty} \right),
\end{equation}
where $\vert\boldsymbol{\nu}\vert_\infty$ is infinity norm of vector $\boldsymbol{\nu}$. Moreover, the following equation holds when $\vert\boldsymbol{\nu}\vert_\infty \in [\beta-\alpha, 2(\beta-\alpha)]$:
\begin{equation}
	1-\frac{\gamma(\beta-\alpha)+\lambda}{\lambda + \gamma\vert \boldsymbol{\nu}\vert_\infty} = \frac{\gamma(\vert \boldsymbol{\nu}\vert_\infty-[\beta-\alpha])}{\lambda + \gamma\vert \boldsymbol{\nu}\vert_\infty} \leq \frac{\gamma(\beta-\alpha)}{\lambda + \gamma\vert \boldsymbol{\nu}\vert_\infty} < \frac{\gamma(\beta-\alpha)}{\lambda}.
\end{equation}
Therefore, we have $\vert G(\mathbf{R}_A)\vert_2 =\gamma(\beta-\alpha)/\lambda$.

We further examine the gap function of the augmented training data $\mathbf{x}+p$, $p\in \mathcal{N}(0, \xi\mathbf{I})$. In this case, the augmented auto-correlation matrix $\mathbf{R}_{A'}$ can be written as
\begin{equation}
	\mathbf{R}_{A'} = \mathbf{R}_A + \xi\mathbf{I},
	\label{eq:aug_z}
\end{equation}
and the augmented gap function is defined as $G(\mathbf{R}_{A'}) = \xi \mathbf{R}^{-1/2}(\mathbf{R}_B-\mathbf{R}_{A'})\mathbf{R}^{-1/2}$.
By substituting Eq. \eqref{eq:aug_z}, we can obtain
\begin{equation}
	\begin{aligned}
		G(\mathbf{R}_{A'}) &= \gamma(\lambda\mathbf{I} +  \gamma\mathbf{U}\mathrm{diag}(\boldsymbol{\nu})\mathbf{U}^\mathsf{T})^{-1/2} [(\beta-\alpha)\mathbf{I} - (\xi\mathbf{I}+ \mathbf{U}\mathrm{diag}(\boldsymbol{\nu})\mathbf{U}^\mathsf{T})](\lambda\mathbf{I} +  \gamma\mathbf{U}\mathrm{diag}(\boldsymbol{\nu})\mathbf{U}^\mathsf{T})^{-1/2} \\
		&= \gamma (\beta-\alpha-\xi + \frac{\lambda}{\gamma}) (\lambda\mathbf{I}+\gamma\mathbf{U}\mathrm{diag}(\boldsymbol{\nu})\mathbf{U}^\mathsf{T})^{-1} -\mathbf{I}.
	\end{aligned}
\end{equation}
Given that $\vert\boldsymbol{\nu}\vert_\infty > \beta-\alpha$, the L2 norm of $G(\mathbf{R}_{A'})$ can be expressed as
\begin{equation}
	\vert G(\mathbf{R}_{A'})\vert_2 = 1-\frac{\gamma(\beta-\alpha -\xi)+ \lambda}{\lambda + \gamma\vert\boldsymbol{\nu}\vert_\infty}.
\end{equation}
Since $\vert\boldsymbol{\nu}\vert_\infty\in [\beta-\alpha, 2(\beta-\alpha)]$. We can draw the following conclusions:
\begin{equation}
	\vert G(\mathbf{R}_{A'})\vert_2 < \frac{\gamma(\beta-\alpha)}{\lambda} = \vert G(\mathbf{R}_{A})\vert_2.
\end{equation}
This conclusion indicates that $\mathbf{R}_{A'}$ is closer to $\mathbf{R}$ than $\mathbf{R}_A$, which proves that data augmentation enables the prediction model to converge to better parameters $\mathbf{w}_{A'}$ compared to $\mathbf{w}_A$.

\bibliographystyle{unsrt}
\bibliography{sample-base}

\end{document}